\documentclass[12pt]{article}
\usepackage{microtype}
\newlength{\defbaselineskip}
\usepackage{algorithmic}
\usepackage{tikz}
\usepackage{caption}
\usepackage{subfigure}
\usepackage{hyperref}
\hypersetup{
     colorlinks   = true,
     linkcolor    = blue,
     citecolor    = green
}
\usepackage{graphicx}
\usepackage{graphics}
\usepackage{multirow}

\usepackage{booktabs} 
\usepackage{epstopdf}
\usepackage[all]{xy}
\usepackage{mathptmx}     
\usepackage{amssymb,mathrsfs,amsmath,amscd,amssymb,bm,color}
\usepackage[mathscr]{euscript}
\usepackage{stackrel}
\usepackage{fancyhdr}
\usepackage{makeidx}
\usepackage{afterpage}
\usepackage{mathtools}
\usepackage{url}
\usepackage{varwidth}
\usepackage[T1]{fontenc}

\DeclareMathOperator*{\argmin}{arg\,min}

\def \u {\mathbf{u}}

\newcommand{\nonl}{\renewcommand{\nl}{\let\nl\oldnl}}% Remove line number for one line

\def\1{\bm{1}}

\newcommand{\R}{\mathbb{R}}

\def\bbu{{\ensuremath{\mathbf u}}}
\def\bbx{{\ensuremath{\mathbf x}}}
\def\bby{{\ensuremath{\mathbf y}}}
\def\bbz{{\ensuremath{\mathbf z}}}
\def\bbm{{\ensuremath{\mathbf m}}}
\def\bbg{{\ensuremath{\mathbf g}}}
\def\bbv{{\ensuremath{\mathbf v}}}
\def\bba{{\ensuremath{\mathbf a}}}
\def\bbb{{\ensuremath{\mathbf b}}}

\def\bbq{{\ensuremath{\mathbf q}}}

\def \R {\mathbb{R}}

\usepackage{amsthm,bm}
\newtheorem{thm}{Theorem}

\newtheorem{rem}[thm]{Remark}
\newtheorem{lem}[thm]{Lemma}

\newtheorem{assum}[thm]{Assumption}

\newcommand{\beq}{\begin{equation}}
\newcommand{\eeq}{\end{equation}}
\newcommand{\beqa}{\begin{eqnarray}}
\newcommand{\eeqa}{\end{eqnarray}}
\newcommand{\beqas}{\begin{eqnarray*}}
\newcommand{\eeqas}{\end{eqnarray*}}

\newtheorem*{lem*}{Lemma}
\usepackage[linesnumbered,ruled,vlined,commentsnumbered]{algorithm2e}

\usepackage{caption}
\usepackage[hang,flushmargin]{footmisc}
\usepackage{lipsum}
\makeatletter
\newcommand{\algorithmfootnote}[2][\footnotesize]{
  \let\old@algocf@finish\@algocf@finish
  \def\@algocf@finish{\old@algocf@finish
    \leavevmode\rlap{\begin{minipage}{\linewidth}
    #1#2
    \end{minipage}}%
  }
}

\begin{document}

\title{Dynamic Regret of Adaptive Gradient Methods for Strongly Convex Problems}
	\vspace{0.8cm}
\author{{Parvin Nazari}\thanks{Department of Mathematics \& Computer Science, Amirkabir University of Technology, Email: \texttt{p$\_$nazari@aut.ac.ir}}
\and{Esmaile Khorram}\thanks{ Department of Mathematics \& Computer Science, Amirkabir University of Technology, Email:  \texttt{eskhor@aut.ac.ir}
}
}
\date{}
	\maketitle
	
\title{}
	\maketitle
Adaptive gradient algorithms such as \textsc{AdaGrad} and its variants have gained popularity in the training of deep neural networks.
While many works as for adaptive methods have focused on the static regret as a performance
metric to achieve a good regret guarantee,
the dynamic regret analyses of these methods remain unclear.
As opposed to the static regret,
dynamic regret is considered to be a stronger concept of performance
measurement in the sense that it explicitly elucidates the non-stationarity of the environment.
In this paper, we go through a variant of \textsc{AdaGrad} (referred to as M-\textsc{AdaGrad} ) in a strong convex setting via the notion of dynamic regret,
which measures the performance of an online learner against a reference (optimal) solution that may change over time.
We demonstrate a regret bound in terms of the path-length of the minimizer sequence that essentially reflects the non-stationarity of environments.
In addition, we enhance the dynamic regret bound by exploiting the multiple accesses of the gradient to the learner in each round.
Empirical results indicate
that M-\textsc{AdaGrad} works also well in practice.
\section{Introduction}

Online convex optimization (OCO) is a fundamental tool for sequential decision making and has found a wide range of applications \cite{hazan2016introduction,hosseini2016online}. The protocol of OCO can be modeled as a repeated game between a learner and an adversary: In each round $t=1,\ldots,T$, the learner picks an action $\bbx_t$ from a convex feasible set $\mathcal{X}$, and at the same time the adversary selects a convex loss function $f_t(\cdot): \mathcal{X} \mapsto \R$, and the
learner incurs an instantaneous loss $f_t(\bbx_t)$. The aim of the learner is to minimize the regret:
\begin{equation} \label{eqn:static-regret}
  {\bf Reg}_T^s(\bbx) = \sum_{t=1}^T f_t(\bbx_t)  - \min_{\bbx \in \mathcal{X}} \sum_{t=1}^T f_t(\bbx),
\end{equation}
which measures the discrepancy between the cumulative loss of the learner and that of the best fixed action in hindsight, and is typically referred to as \emph{static} regret since the comparator is time-invariant. The classical online gradient descent (OGD) enjoys $O(\sqrt{T})$ and $O(\log T)$ upper bound on static regret
for convex and strongly convex functions, respectively \cite{zinkevich2003online, hazan2007logarithmic}.

Although static regret has been extensively studied, when the environment is changing, its performance is no longer suitable since the time-invariant comparator in (\ref{eqn:static-regret}) may behave badly.
To circumvent this difficulty, recent studies have introduced new forms of performance metric, including dynamic regret.
The dynamic regret is defined as the difference between the cumulative loss of the learner and that of a sequence of comparators $\u_1, \ldots, \u_T \in \mathcal{X}$ \cite{zinkevich2003online}:
\begin{equation*}
{\bf Reg}_T^d(\u_1,\ldots,\u_T) = \sum_{t=1}^T f_t(\bbx_t)  - \sum_{t=1}^T f_t(\u_t).
\end{equation*}
Dynamic regret  measures the learner's performance in the sense that the comparator changes over time.
The concept of dynamic regret is interesting in many applications, say online
recommendation (since the customers' preference always evolves over time).
Most studies on dynamic regret only compare the cumulative loss of the learner against a sequence of  minimizers of the loss functions
\cite{jadbabaie2015online,besbes2015non,yang2016tracking,mokhtari2016online,zhang2017improved,nazari2021dynamic}:
\begin{equation} \label{eqn:dynamic:2}
\begin{split}
 {\bf Reg}_T^d(\bbx^*_1,\ldots,\bbx^*_T) =\sum_{t=1}^T f_t(\bbx_t)  - \sum_{t=1}^T   f_t(\bbx_t^*)=\sum_{t=1}^T f_t(\bbx_t)  -\sum_{t=1}^T \min_{\bbx \in \mathcal{X}} f_t(\bbx),
\end{split}
\end{equation}
where $\bbx_t^* := \argmin_{\bbx \in \mathcal{X}} f_t(\bbx)$ is a minimizer of $f_t(\cdot)$ over domain $\mathcal{X}$. It is well-known that a sublinear dynamic regret is unattainable in the worst case, unless we
impose some certain regularities of the comparator sequence or the function sequence.
For example, \cite{zinkevich2003online} demonstrates that OGD with a constant stepsize enjoys $O(\sqrt{T} D_T)$ dynamic regret bound,
 where $D_T$ is the path-length of the comparator sequence as
\begin{equation*}
D_T:=D(\bbu_{1},\ldots,\bbu_{T})=\sum_{t=2}^{T}\|\bbu_{t}-\bbu_{t-1}\|.
 \end{equation*}
This upper bound is adaptive in the sense that it automatically becomes tighter when the comparators change slowly.
In a follow-up work, \cite{hall2013dynamical} put forward a variant of path-length
\[
D_{\Phi,T}:=D_{\Phi}(\u_1,\ldots,\u_T)=\sum_{t=2}^T \|\u_{t} - \Phi_{t} (\u_{t-1})\|,
\]
where $\Phi_t (\cdot):\mathcal{X} \mapsto \mathcal{X}$ stands for the predicted comparator point for the $t^{th}$ round.
Then, they deployed a new
method, dynamic mirror descent, which enjoys an $O(\sqrt{T} D_{\Phi,T})$ dynamic regret.
Whenever the
comparator points are the optimal points in \eqref{eqn:dynamic:2}, i.e., $\u_t=\bbx_t^*$, a natural regularity defined as
\begin{equation}\label{CT}
D_T^*:=D(\bbx^*_{1},\ldots,\bbx^*_{T})=\sum_{t=2}^{T}\|\bbx^*_{t}-\bbx^*_{t-1}\|.
 \end{equation}
Toward strongly convex and smooth functions, or convex and smooth functions provided
that all the minimizers lie in the interior of $\mathcal{X}$, OGD has been investigated in \cite{mokhtari2016online} and \cite{yang2016tracking} to achieve a regret of $O(D^*_T)$.
Another regularity of the comparator sequence is the squared
path-length
 \begin{equation}\label{CT2}
  S_T^*:=S(\bbx^*_{1},\ldots,\bbx^*_{T})=\sum_{t=2}^{T}\|\bbx^*_{t}-\bbx^*_{t-1}\|^2,
\end{equation}
which could be smaller than the path-length $D_T^*$ when local minimizers move slowly.
In a subsequent work, \cite{zhang2017improved} suggested online multiple gradient descent (OMGD) method, and argued an $O(\min \{D_T^*, S_T^*\})$ regret bound for the case of (semi-)strongly convex and smooth functions.

Apart from the path-length of the comparator sequence, recent work in \cite{besbes2015non} have proposed the functional variation
\begin{equation*}
  V_T:=V(f_1,\ldots,f_T)=\sum_{t=2}^{T} \max_{\bbx\in \mathcal{X}}|f_t(\bbx)-f_{t-1}(\bbx)|
\end{equation*}
to evaluate the performance of OGD for the case when a noisy estimate of the gradient is available.
They revealed that applying restarted online gradient descent method brings about upper bounds of $O(T^{2/3}(M_T+1)^{1/3})$ and $O(\log T\sqrt{T(M_T+1)})$ on dynamic regret under the assumption that $V_T\leq M_T$ in which $M_T$ is given beforehand for convex and strongly convex functions, respectively.
For convex and smooth functions, \cite{chiang2012online} indicated an $O(\sqrt{G_T})$ regret bound, where
$G_T:=G(f_1,\ldots,f_T)=\sum_{t=2}^{T} \max_{\bbx\in \mathcal{X}}\|\nabla f_t(\bbx)-\nabla f_{t-1}(\bbx)\|^2$ is the gradient variation. Gradient-variation bounds are
particularly favored in slowly changing environments in which the online functions evolve
gradually.

The idea of adapting first order optimization methods to online convex learning have been studied both theoretically and empirically for convex \cite{duchi2011adaptive,kingma2014adam,reddi2019convergence}, strongly convex \cite{chen2018sadagrad,wang2019sadam}, and non-smooth non-convex settings \cite{nazari2020adaptive}.
\textsc{AdaGrad} was postulated by \cite{duchi2011adaptive}, which works well for sparse gradients as it invokes all the past gradients
to scale the gradient.
In a subsequent work, \cite{mukkamala2017variants} put forth \textsc{Sc-Adagrad}, which enjoyed logarithmic regret bounds for strongly convex functions.
\cite{chen2018sadagrad} suggested \textsc{Sadagrad} for solving stochastic strongly convex optimization and more generally stochastic convex optimization that satisfies the second order growth condition.
The most popular variant of Adagrad is Adam \cite{kingma2014adam}.
The static regret for a variant of \textsc{Adam}
(referred to as \textsc{SAdam} in the strong convex setting) was demonstrated in \cite{wang2019sadam} with a data-dependent $O(\log T)$ convergence rate, but, to our knowledge, no analysis driven by dynamic regret has ever been published. Table~\ref{tab1} summarizes the existing literature on regret bounds.

\begin{table}[t]
\centering
\hfill
\renewcommand{\arraystretch}{1.2}
\begin{minipage}[b]{0.95\linewidth}
\resizebox{\columnwidth}{!}{
\begin{tabular}{lllllll}
\toprule[1.5pt]
Reference & Setting & Problem Type &Regret Bound& Algorithm & Adaptive & Momentum
\\
\cline{1-7}
\cite{duchi2011adaptive}           &Static               &   Convex                    &  $O\left(\sqrt{T}\right)$            & \textsc{AdaGrad}     & Yes & No\\
\cite{mukkamala2017variants}       & Static              &   Strongly convex           &  $O\left( \log T\right)$             & SC-\textsc{AdaGrad}  & Yes & No \\
\cite{nazari2019dadam} \qquad      &Dynamic              &  Convex                     & $O\left( \sqrt{T }{C}^*_T\right)$    &\textsc{DAdam}        &Yes & Yes\\
\cite{zhang2017improved} \qquad    & Dynamic             &  Strongly Convex and Smooth &$O\left(\min\{ {D}^*_T,S^*_T\}\right)$& OMGD                 & No &No \\
This paper \qquad                  &Dynamic              &  Strongly Convex and Smooth & $O\left( {C}^*_T\right)$             &M-\textsc{AdaGrad}    &Yes& Yes\\
This paper \qquad                  &Dynamic              &  Strongly Convex and Smooth &
$O\left(\min\{ {C}^*_T,S^*_T\}\right)$&  MM-\textsc{AdaGrad} &Yes& Yes\\
\cline{1-7}
\bottomrule[1.5pt]
\end{tabular}}
\caption{Summary of the regret bounds in this paper and comparison with regret bounds in the literature. The complexity measures $D_T^*$, $ {S}^*_{T}$ and $C_T^*$ are defined in \eqref{CT}, \eqref{CT2} and \eqref{CTT}, respectively.}
\label{tab1}
\end{minipage}

\end{table}
\paragraph{Contributions.}
In this paper, we carry out the dynamic regret analysis of \textsc{AdaGrad} equipped with momentum (of parameter $\beta$), under the strong-convexity and smoothness condition of the objective function.
We present Momentum \textsc{AdaGrad} (M-\textsc{AdaGrad}) in accordance with the basic \textsc{AdaGrad}, while the direction is constructed by means of an exponential moving average of the past gradients.
We allude that M-\textsc{AdaGrad} achieves the dynamic regret bound of $O(C^*_T)$ with one gradient query in each round.
Inspired by the online multiple gradient descent methods developed by \cite{zhang2017improved}, where multiple gradients are accessible to the learner in one round, we bring up MM-\textsc{AdaGrad} to obtain
a tighter dynamic regret bound of order $O(\min\{C^*_T,S^*_T\})$, provided the gradients of minimizers are small.
In this way, if the local variations ($\|\bbx_t^*-\bbx_{t-1}^* \|$'s) are small, ${S}^*_{T}$ can be remarkably smaller than ${C}^*_T$.

\paragraph{Outline.}
The remainder of the paper is organized as follows. In Section~\ref{bc}, we record some basic notation and expound the concept of \textsc{Adagrad} method. Section~\ref{sec:main} presents an analysis framework and main results.
In section~\ref{sec:main2}, we propose our MM-\textsc{AdaGrad} algorithm and upper bound the dynamic regret for strongly convex and smooth functions.
Section \ref{exper} presents numerical experiments.
Finally, Section~\ref{conclusion} concludes this paper. The detailed proofs of the main results established are delegated to the Appendix.

\section{Mathematical Preliminaries and Notations.}\label{bc}
\subsection{Notations}\label{sec:notation}
Throughout, $\mathbb{R}_+$ and $\mathbb{R}^p$ denote the sets of nonnegative real numbers and real coordinate space of $p>0$ dimensions, respectively.
Vectors are denoted by lower case bold face letters, scalars by lower case letters, and matrices by upper case letters.
For any vectors $\bba,\bbb\in\mathbb{R}^p$, we use $\bba^{1/2}$ to denote element-wise square root, $\bba^2$ to denote element-wise square, $\bba/\bbb$ to denote element-wise division, $ \langle \bba , \bbb \rangle$ and $\bba \odot \bbb$ indicate standard Euclidean inner product and element-wise product, respectively.
$\mathbf{1}_p$ is used to denote the $p$ vector of ones.
For a positive integer $n\in \mathbb{N}$, we set $[n]:=\{1,\ldots,n\}$. Further, for any vector $\bbx_t\in \mathbb{R}^p$, $x_{t,i}$ denotes its $i^{th}$ coordinate where $i\in[p]$.
We also denote $x_{1:t,i}=[x_{1,i},x_{2,i},\ldots,x_{t,i}]^\top$.
We let $\text{diag}(\bbx)$ denote the diagonal matrix with diagonal entries $x_1, \ldots , x_p$.
We use $\|\cdot\|$, $\|\cdot\|_1$ and $\|\cdot\|_{\infty}$ to denote the $\ell_2$-norm, $\ell_1$-norm and the infinity norm, respectively. For $\upsilon_i>0$, $\forall i\in [p]$, we define a weighted norm $\|\bbx \|^2_{\bbv}:=\langle \bbx,(\text{diag}\, \bbv) \bbx\rangle$ and a weighted projection operator onto $\mathcal{X}$:
\begin{equation*}
\Pi_{\mathcal{X},\bbv}~\big(\bbx\big)=\argmin_{\bby\in \mathcal{X}} \|\bbx-\bby\|_{\bbv}^2.
\end{equation*}
A continuously
differentiable function $f:\mathcal{X}\mapsto \mathbb{R}$ is said to be $\lambda$-strongly convex with respect to weighted norm $\|\cdot \|_{\bbv}$ if and only if for some $\lambda \in \R_{+}$,
$
f(\bby)\geq f(\bbx)+\langle \nabla f(\bbx),\bby-\bbx\rangle+\frac{\lambda}{2}\|\bby-\bbx\|^2_{\bbv}, ~~ \forall  ~~\bbx,\bby \in \mathcal{X}
$.
Also, it is said to be $L$-smooth if for some $L \in \R_{+}$,
$
f(\bby)\leq f(\bbx)+\langle \nabla f(\bbx),\bby-\bbx\rangle+\frac{L}{2}\|\bby-\bbx\|^2_{\bbv}, ~~ \forall  ~~\bbx,\bby \in \mathcal{X}.
$
\subsection{Review of \textsc{Adagrad} Method}
Variants of adaptive gradient methods have been put forward to adjust automatically the learning rate by virtue of some forms of the
past gradients to scale coordinates of the gradient. \textsc{Adagrad}~\cite{duchi2011adaptive} is the first popular
method in this line with adaptive learning rate for each individual dimension, which in turn
is thought to be effective for sparse optimization.
In detail, \textsc{Adagrad} adopts the following
update form:
\begin{equation*}
	\bbx_{t+1}=\bbx_t-\alpha {\bbv}_t^{-1/2} \bbg_t,\quad {\bbv}_t=\frac{1}{t}\sum_{j=1}^{t}\bbg_j\odot \bbg_j,
\end{equation*}
where $\alpha>0$, $\bbg_t=\nabla f_t(\bbx_t)$. When $\bbv_t=\mathbf{1}_p$, \textsc{AdaGrad}  reduces to SGD, which scales the gradient uniformly in all dimensions. As compared to SGD, \textsc{AdaGrad} dynamically incorporates knowledge of history gradients to carry out more informative gradient-based learning.
As a result, larger learning rates are performed for components with smaller gradients, while smaller learning rates are performed for components with larger gradients, resulting in faster convergence than SGD for sparse gradients in the approach.
Various variations of \textsc{AdaGrad} such as \textsc{RMSProp} \cite{tieleman2017divide}, \textsc{Adadelta} \cite{zeiler2012adadelta},\textsc{Adam} \cite{kingma2014adam}, \textsc{AMSGrad}~\cite{reddi2019convergence}, \textsc{Dadam} \cite{nazari2019dadam} and so on have been proposed so as to boost the performance of it.
\textsc{AdaGrad} attains  the  well-known  data-dependent  regret  bound $O\left(\sum_{i=1}^{p}\|g_{1:T,i}\|\right)$, where $T$ is the iteration number and $g_{1:T,i}$ is a vector of historical stochastic  gradients of the $i^{th}$ dimension to train online convex problems.
The data-dependent regret  bound  outperforms the  original $O(\sqrt{T})$ regret bound,   which   is   identified as  optimal   \cite{hazan2019introduction},   when   the gradients  are  sparse  or  very  small.
The results of \cite{duchi2011adaptive} further reflect the convergence of \textsc{Adagrad} with the rate of $O(G_{\infty}^2\sum_{i=1}^{p}\log(\|g_{1:T,i}\|))$
for strong convex settings where $G_{\infty}$ represents the upper bound of stochastic gradient's infinity norm.
Furthermore, this dependence on $\|g_{1:T,i}\|$ has been  also presented in two recent variants of \textsc{AdaGrad} adapted to the strongly convex case, namely, SC-\textsc{AdaGrad} \cite{mukkamala2017variants} and \textsc{MetaGrad} \cite{van2016metagrad}.
\begin{algorithm}[t]
\caption{Momentum AdaGrad (M-\textsc{AdaGrad}).}\label{alg:subgradient}
\SetKwInput{Input}{input~}
\SetKwInput{Output}{output~}
\SetKwInput{optionone}{option~I~}
\SetKwInput{optiontwo}{option~II~}
\Input{ Initial point $\bbx_1 \in \mathcal{X}$, number of iterations $T$,  stepsize $\alpha$ and decay parameter $\beta<1$. \\}
\BlankLine
Initialize $\bbm_{0}=\bbv_{0}=0 .$\\
\For{ $1\leftarrow t : T$ }{
$\bbg_t = \nabla f_t(\bbx_t)$.\\
$\bbm_{t}= \beta \bbm_{t-1}+(1-\beta)\bbg_{t}$.\\ 
$\bbv_{t}= \bbv_{t-1}+ \bbg_{t}\odot \bbg_{t} $.\\
 $ \bbx_{t+1}=\Pi_{\mathcal{X}, {\bbv}^{1/2}_t}\big(\bbx_{t} - \alpha{\bbv}_t^{-1/2} \bbm_t\big)$.
}
Return {$\bbx_T$}
\end{algorithm}
\section{Momentum AdaGrad}\label{sec:main}
In this section, we provide the M-\textsc{AdaGrad} algorithm and its theoretical analysis.
The pseudocode for Momentum \textsc{AdaGrad} (M-\textsc{AdaGrad}) is formally described in Algorithm \ref{alg:subgradient} integrating \textsc{AdaGrad} with the direction $\bbm_t$ which is the exponential moving average of the gradients used in the paper by \cite{kingma2014adam}. Here the square root, the square, and the division operators are taken elementwise.
The standard regret analysis in \cite{kingma2014adam} assumes a fast diminishing schedule for $\beta$ parameter, while \cite{alacaoglu2020new} obviates this requirement and leverages a constant $\beta$ parameter in Adam-type methods.
Our analysis renders the first step toward understanding adaptive learning rate methods in the strongly convex
and smooth setting where  environments  change  over time.
In order to proceed, it is necessary to postulate some standard conditions for the regret analysis.

\begin{assum}\label{as1}
The loss functions $f_t$ for all $t\in[T]$ are $\lambda$-strongly convex and $L$-smooth over $\mathcal{X}.$
\end{assum}
\begin{assum}\label{as2}
The infinite norm of the gradients of all loss functions is bounded by $G_{\infty}$, i.e.,
for all $\bbx\in \mathcal{X}$ and $t\in [T]$, it holds that
$ \|\nabla f_t(\bbx) \|_{\infty}\leq G_{\infty}.$
\end{assum}

\begin{assum}\label{as3}
The set $\mathcal{X}$ is convex and compact with diameter
$D_{\infty}$, i.e., for all $\bbx,\bby\in \mathcal{X}$, it holds that $\|\bbx-\bby\|_{\infty}\leq D_{\infty}.$
\end{assum}
The strong convexity of functions $f_t$ in Assumption~\ref{as1} guarantees the existence of a unique minimizer $\bbx_t^*$ for the function $f_t$ over the convex set $\mathcal{X}$, and
the smoothness of $f_t$ is the standard assumption.
In particular, it is worth noting that Assumption \ref{as2} is slightly weaker than the $\ell_2$-boundedness assumption $\|\nabla f_t(\bbx) \|\leq G_2$ used in \cite{mokhtari2016online,zhang2017improved}.
Owing to $\|\nabla f_t(\bbx) \|_{\infty}\leq \|\nabla f_t(\bbx) \|$, the $\ell_2$-boundedness assumption leads to Assumption \ref{as2} with $G_{\infty} = G_2$.
In fact, $G_2$ is often larger than $G_{\infty}$ by a factor of $\sqrt{p}$.
The Assumption~\ref{as3} has already been used by many authors \cite{reddi2018convergence,nazari2019dadam,nazari2019adaptive}
to establish the convergence of adaptive gradient methods. We will consider the following $\ell_1$-based regularity measure
\begin{equation}\label{CTT}
C_{T,i}^*:= \sum_{t=1}^{T-1} |x^*_{t+1,i}-x^*_{t,i}|,  \quad \text{for all} \quad  i \in [p],
\end{equation}
which captures the cumulative difference between successive
comparators $\{\bbx^{*}_t\}_{t=1}^{T}$. Throughout, we set $C_{T}^*:=\sum_{i=1}^p C_{T,i}^*$.

Our main result (Theorem~\ref{Sthm:stochastic_sub}) establishes a bound
on the dynamic regret ${\bf Reg}_T^d(\bbx^*_1,\ldots,\bbx^*_T)$ in the sense of \eqref{eqn:dynamic:2} in terms of $C^*_T$.
Having stated the theorem, we then display that under mild conditions, our results recover previous rates on online gradient descent in dynamic setting when the function is strongly convex and smooth.
To prove Theorem~\ref{Sthm:stochastic_sub}, we start by presenting the following lemma which stipulates an upper bound on the distance between
an action $\bbx_{t+1}$ and the optimal argument $\bbx_t^*$.
\begin{lem} \label{lem:strong:convex}
Suppose that Assumptions \ref{as1}-\ref{as3} hold and let $\alpha\leq 1/L$.
Then, the decision sequence $\{\bbx_{t+1}\}$ generated by Algorithm~\ref{alg:subgradient} satisfies
\begin{align}\label{fff}
& \sum_{t=1}^{T} \sum_{i=1}^p {\upsilon}^{1/2}_{t,i}(x_{t+1,i}- x_{t,i}^*)^2
\leq
\bar{\sigma}\sum_{t=1}^{T} \sum_{i=1}^p {\upsilon}^{1/2}_{t,i}(x_{t,i}- x_{t,i}^*)^2
+\frac{\beta \tilde{\sigma}\vartheta}{1-\beta},
\end{align}
where $\bbx_t^* =\argmin_{\bbx \in \mathcal{X}} f_t(\bbx)$, ${C}^*_{T,i}$ is defined as in \eqref{CTT}, and
\begin{align*}
{\bar{\sigma}}&:=1-\frac{2\lambda}{\frac{1}{\alpha}+\lambda},\,\,\,\,
{\tilde{\sigma}}:=\frac{2}{\lambda+\frac{1}{\alpha}},
\\\vartheta&:=\sum_{i=1}^{p}\big(\frac{ D_{\infty}^2}{2\alpha}+\beta {C}^*_{T,i}\big) {\upsilon}_{T,i}^{1/2}+\frac{2\alpha}{(1 - \beta)^2} \sum_{i=1}^p \|g_{1:T,i} \|.
\end{align*}
\end{lem}
The proof is deferred to the appendix.

\begin{rem}
The result in Lemma~\ref{lem:strong:convex} provides an upper bound on the cumulative squared distance between
$x_{t+1,i}$ and the current optimal $x_{t,i}^*$ over $T$ time instances.
Different
from Proposition 2 in \cite{mokhtari2016online} and Lemma 5 in \cite{zhang2017improved}, Lemma \ref{lem:strong:convex}
takes into account the impacts of adaptive gradient methods and recovers the results in \cite{mokhtari2016online} and \cite{zhang2017improved} as special cases.
\end{rem}

With the result of Lemma \ref{lem:strong:convex} in hand, we can now characterize for the first time dynamic regret analysis of M-\textsc{AdaGrad} algorithm.

\begin{thm}\label{Sthm:stochastic_sub}
Suppose that Assumptions \ref{as1}-\ref{as3} hold and let $\alpha\leq 1 / L$.
Then, the dynamic regret \eqref{eqn:dynamic:2} achieved by Algorithm~\ref{alg:subgradient} satisfies
\begin{align}\label{eqn:m-adagrad:bound}
 {\bf Reg}_T^d(\bbx^*_1,\ldots,\bbx^*_T)&\leq  \varpi_1\sum_{i=1}^{p}\|g_{1:T,i} \|+\varpi_2\sum_{i=1}^{p}\upsilon_{T,i}^{1/2}{C}^*_{T,i}+\varpi_3,
\end{align}
where $\bbx_t^* =\argmin_{\bbx \in \mathcal{X}} f_t(\bbx)$, $\bar{\sigma}$, $\tilde{\sigma}$, ${C}^*_{T,i}$ are defined as in Lemma \ref{lem:strong:convex}, and
\begin{align*}
\nonumber\varpi_1&:=1+\big(1-{\bar{\sigma}}\big)^{-1}\frac{\alpha \beta \tilde{\sigma}}{(1-\beta)^3},
\qquad \,\, \varpi_2:=\frac{1}{2}\big(1-{\bar{\sigma}}\big)^{-1}\big(\frac{\tilde{\sigma}\beta^2}{1-\beta}+2D_{\infty} \big),
\\\varpi_3&:=\frac{1}{2}\big(1-{\bar{\sigma}}\big)^{-1}\Big(\sum_{i=1}^p {\upsilon}^{1/2}_{1,i}(x_{1,i}- x_{1,i}^*)^2
+\big( \frac{\beta \tilde{\sigma} }{2(1-\beta)\alpha} +1\big) D_{\infty}^2\sum_{i=1}^p  {\upsilon}^{1/2}_{T,i}\Big).
\end{align*}
\end{thm}
\begin{proof}
According to the Mean Value Theorem, there exists a vector $\mathbf{w}\in\{\mathbf{y}|\mathbf{y}=\delta \bbx_{t}+(1-\delta)\bbx_{t}^*,\delta \in [0,1] \}$ such that
\begin{align} \label{eqn:thm1:11}
\nonumber {\bf Reg}_T^d(\bbx^*_1,\ldots,\bbx^*_T)&=\sum_{t=1}^T \big(f_{t}(\bbx_{t}) - f_{t}(\bbx_{t}^*)\big)
\\\nonumber&\leq \sum_{t=1}^T\langle \nabla f_t(\mathbf{w}),\bbx_t-\bbx_{t}^* \rangle
\\ \nonumber &\leq \frac{1}{2}\sum_{t=1}^T\|\nabla f_t(\mathbf{w})\|_{\bbv_t^{-1/2}}^2+\frac{1}{2}\sum_{t=1}^T\|\bbx_{t} - \bbx_{t}^*\|_{\bbv_t^{1/2}}^2
\\&\leq \sum_{i=1}^{p}\|g_{1:T,i} \|+\frac{1}{2}\sum_{t=1}^T\|\bbx_{t} - \bbx_{t}^*\|_{\bbv_t^{1/2}}^2,
\end{align}
where the second inequality is due to $2\langle a,b \rangle\leq \| a\|^2+\|b\|^2$, the third inequality is by Lemma~\ref{lem:simple-grad-boundd}.
We next bound $\sum_{t=1}^T \|\bbx_{t} - \bbx_{t}^*\|_{\bbv_t^{1/2}}^2$. To this end, we observe the following:
\begin{align}\label{nkj}
\nonumber  &\quad\sum_{t=1}^T \sum_{i=1}^p  \upsilon_{t,i}^{1/2}(x_{t,i} - x_{t,i}^*)^2 \\\nonumber&=\sum_{t=1}^T \sum_{i=1}^p\big( \upsilon_{t,i}^{1/2}(x_{t+1,i} - x_{t+1,i}^*)^2-\upsilon_{t,i}^{1/2}(x_{t+1,i} - x_{t,i}^*)^2\big)
\\\nonumber&+\sum_{t=1}^T \sum_{i=1}^p\big( \upsilon_{t,i}^{1/2}(x_{t,i} - x_{t,i}^*)^2-\upsilon_{t,i}^{1/2}(x_{t+1,i} - x_{t+1,i}^*)^2\big)
\\
&+\sum_{t=1}^T \sum_{i=1}^p \upsilon_{t,i}^{1/2}(x_{t+1,i} - x_{t,i}^*)^2 .
\end{align}
We proceed to upper bound each of the three terms
on the right-hand side of \eqref{nkj}. Assumption~\ref{as3} implies
\begin{subequations}

\begin{align}\label{re}
\nonumber &\quad \sum_{t=1}^T \sum_{i=1}^p\big(\upsilon_{t,i}^{1/2}(x_{t+1,i} - x_{t+1,i}^*)^2-\upsilon_{t,i}^{1/2} (x_{t+1,i} - x_{t,i}^*)^2\big)
\\\nonumber &=\sum_{t=1}^T \sum_{i=1}^p \upsilon_{t,i}^{1/2}| x^*_{t+1,i}-2x_{t+1,i} + x^*_{t,i} ||x^*_{t,i}-x^*_{t+1,i}   | \\\nonumber
&\leq\sum_{t=1}^T \sum_{i=1}^p \upsilon_{t,i}^{1/2}\big(| x^*_{t+1,i}-x_{t+1,i} |+| x^*_{t,i}-x_{t+1,i}  |\big)| x^*_{t,i}-x^*_{t+1,i}   |\\\nonumber
& \leq 2D_{\infty}\sum_{t=1}^T\sum_{i=1}^p \upsilon_{t,i}^{1/2}|x^*_{t+1,i} - x^*_{t,i} |
\\&\leq 2D_{\infty}\sum_{i=1}^{p}\upsilon_{T,i}^{1/2}{C}^*_{T,i} ,
\end{align}
where ${C}^*_{T,i}$ is defined as in \eqref{CTT}.
Moreover, from Assumption~\ref{as3}, we have
\begin{align}\label{weq1}
\nonumber &\quad \sum_{t=1}^T \sum_{i=1}^p\big( \upsilon_{t,i}^{1/2}(x_{t,i} - x_{t,i}^*)^2-\upsilon_{t,i}^{1/2}(x_{t+1,i} - x_{t+1,i}^*)^2\big)
\\\nonumber &= \sum_{i=1}^p \upsilon_{1,i}^{1/2}(x_{1,i} - x_{1,i}^*)^2+\sum_{t=2}^T\sum_{i=1}^p({\upsilon}^{1/2}_{t,i}-{\upsilon}^{1/2}_{t-1,i})(x_{t,i} - x_{t,i}^*)^2
\\ &\leq
\sum_{i=1}^p \upsilon_{1,i}^{1/2}(x_{1,i} - x_{1,i}^*)^2
+ D_{\infty}^2\sum_{i=1}^p  {\upsilon}^{1/2}_{T,i}.
\end{align}
\end{subequations}
Substituting \eqref{re} and \eqref{weq1} into \eqref{nkj} and using Lemma~\ref{lem:strong:convex}, we then obtain
\begin{align*}
\nonumber\sum_{t=1}^T \sum_{i=1}^p  {\upsilon}^{1/2}_{t,i}(x_{t,i} - x_{t,i}^*)^2
 &\leq\sum_{i=1}^p  {\upsilon}^{1/2}_{1,i}(x_{1,i}- x_{1,i}^*)^2 +{\bar{\sigma}}\sum_{t=1}^{T} \sum_{i=1}^p {\upsilon}^{1/2}_{t,i} (x_{t,i}- x_{t,i}^*)^2
 \\\nonumber&+\frac{\beta  \tilde{\sigma}\vartheta}{1-\beta}+ D_{\infty}^2\sum_{i=1}^p  {\upsilon}^{1/2}_{T,i}+2D_{\infty}\sum_{i=1}^{p}\upsilon_{T,i}^{1/2}{C}^*_{T,i}.
\end{align*}
Rearranging the terms, we get
\begin{align}\label{plk}
\nonumber \sum_{t=1}^T \sum_{i=1}^p  {\upsilon}^{1/2}_{t,i}(x_{t,i} - x_{t,i}^*)^2
&\leq \big(1-{\bar{\sigma}}\big)^{-1}\Big(\sum_{i=1}^p  {\upsilon}^{1/2}_{1,i}(x_{1,i}- x_{1,i}^*)^2
\\&+\frac{\beta  \tilde{\sigma}\vartheta}{1-\beta}+ D_{\infty}^2\sum_{i=1}^p  {\upsilon}^{1/2}_{T,i}+2D_{\infty}\sum_{i=1}^{p}\upsilon_{T,i}^{1/2}{C}^*_{T,i}\Big).
\end{align}
Combining \eqref{plk} and \eqref{eqn:thm1:11}, we get
\begin{align*}
\nonumber\sum_{t=1}^T \big(f_{t}(\bbx_{t}) - f_{t}(\bbx_{t}^*)\big)
&\leq \sum_{i=1}^{p}\|g_{1:T,i} \|+\frac{1}{2}\big(1-{\bar{\sigma}}\big)^{-1}\Big(\sum_{i=1}^p  {\upsilon}^{1/2}_{1,i}(x_{1,i}- x_{1,i}^*)^2
\\&+\frac{\beta {\tilde{\sigma}}\vartheta}{1-\beta}+ D_{\infty}^2\sum_{i=1}^p  {\upsilon}^{1/2}_{T,i}+2D_{\infty}\sum_{i=1}^{p}\upsilon_{T,i}^{1/2}{C}^*_{T,i} \Big).
\end{align*}
This completes the proof.
\end{proof}
\begin{rem}\label{rem:sparsity}
Note that when the data features are sparse and have bounded gradients we obtain
$\sum_{i=1}^p \| g_{1:T,i}\|  \ll  pG_{\infty} \sqrt{T}
$ and
$
\sum_{i=1}^p \upsilon_{T,i}^{1/2} \ll	 p G_{\infty}.
$
This implies that the summation terms in \eqref{eqn:m-adagrad:bound} can be much smaller than their upper bounds for functions with sparse gradients. Thus, similar to adaptive methods,  \textsc{M-AdaGrad} and \textsc{MM-AdaGrad} can achieve significantly better regret bounds compared to vanilla online gradient-type methods \cite{mokhtari2016online} and \cite{zhang2017improved}, respectively.
\end{rem}
\section{Multiple Momentum \textsc{AdaGrad}}\label{sec:main2}
In this section, we come up how to boost the dynamic regret via enabling the learner to query the gradient multiple times.
By taking into account the possibility that a learner may access the gradient of a function multiple time, we show the proposed algorithm MM-\textsc{AdaGrad} in Algorithm \ref{alg:subgradient2}. As such, we are able to draw more information from each function and thus are more likely to gain a tight bound on the dynamic regret. We consider the bounded variability of the reference sequence in terms of
\begin{equation}\label{STT}
S_{T,i}^*:= \sum_{t=1}^{T-1} (x^*_{t+1,i}-x^*_{t,i})^2,  \quad \text{for all} \quad  i \in [p],
\end{equation}
which captures the cumulative difference between successive
comparators $\{\bbx^{*}_t\}_{t=1}^{T}$. Throughout, we set $S_{T}^*:=\sum_{i=1}^p S_{T,i}^*$.
In the proof of our main theorem, we make use of the technical lemma provided below.
\begin{algorithm}[t]
\caption{Multiple Momentum \textsc{AdaGrad} (MM-\textsc{AdaGrad}).}\label{alg:subgradient2}
\SetKwInput{Input}{input~}
\SetKwInput{Output}{output~}
\SetKwInput{optionone}{option~I~}
\SetKwInput{optiontwo}{option~II~}
\Input{ Initial point $\bbx_1 \in \mathcal{X}$, number of iterations $T$, number of inner iterations $K$, stepsize $\alpha$ and decay parameter $\beta<1$. \\}
\BlankLine
Initialize $\bbm_{0}=\bbv_{0}=0 $.\\
\For{ $1\leftarrow t : T$ }{\label{forins}
$\bbz_t^1=\bbx_t$.\\
\For{ $1\leftarrow j : K$ }{
$\bbg^j_t = \nabla f_t(\bbz_t^j)$.\\
$\bbm^j_{t}= \beta \bbm^{j}_{t-1}+(1-\beta)\bbg^j_{t}$.\\
$\bbv^j_{t}= \bbv^{j}_{t-1}+ \bbg^j_{t} \odot \bbg^j_{t}$.\\
$ \bbz_{t}^{j+1}=\Pi_{\mathcal{X}, ({\bbv}^j_t)^{1/2}}\big(\bbz_{t}^j - \alpha({\bbv}^j_t)^{-1/2} \bbm^j_t\big)$.}
 $ \bbx_{t+1}=\bbz_{t}^{K+1}$.\label{m11}
}
Return {$\bbx_T$}
\end{algorithm}

\begin{lem}\label{lem:strong:convex2}
Suppose that Assumptions \ref{as1}-\ref{as3} hold and let $\alpha\leq 1/L$.
Then, the decision sequence $\{\bbz^{j+1}_{t}\}$ generated by Algorithm~\ref{alg:subgradient2} satisfies
\begin{align*}
\nonumber& \sum_{t=1}^{T} \sum_{i=1}^p {({\upsilon}^{j}_{t,i})^{1/2}}(z^{j+1}_{t,i}- x_{t,i}^*)^2
\leq
{\bar{\sigma}}\sum_{t=1}^{T} \sum_{i=1}^p {({\upsilon}^{j}_{t,i})^{1/2}}(z^j_{t,i}- x_{t,i}^*)^2+\frac{\beta {\tilde{\sigma}}\vartheta}{1-\beta},
\end{align*}
where
$\bbx_t^* =\argmin_{\bbx \in \mathcal{X}} f_t(\bbx)$,
 ${S}^*_{T,i}$ is defined as in \eqref{STT}, and
 \begin{align}\label{kkk}
\nonumber&{\bar{\sigma}}:=1-\frac{2\lambda}{\frac{1}{\alpha}+\lambda},\quad \quad
{\tilde{\sigma}}:=\frac{2}{\lambda+\frac{1}{\alpha}},\\
&\vartheta:=\sum_{i=1}^{p}\big(\frac{ D_{\infty}^2}{2\alpha}+\frac{S^*_{T,i}}{2}\big) ({\upsilon}_{T,i}^j)^{1/2}
+\frac{2\alpha}{(1 - \beta)^2} \sum_{i=1}^p \|g^j_{1:T,i} \|.
\end{align}
\end{lem}
Equipped with this lemma, we now state our main result.
The subsequent theorem shows that the multiple accesses to the gradient do indeed help to improve the dynamic regret.
\begin{thm}\label{momo2}
Suppose that Assumptions \ref{as1}-\ref{as3} hold . Let $\alpha\leq 1/L$ and $K=\lceil \frac{1/\alpha+\lambda}{2\lambda}\ln 4\rceil$. Then, for any constant $\gamma>0$, the dynamic regret \eqref{eqn:dynamic:2} achieved by Algorithm~\ref{alg:subgradient2} satisfies

\begin{align}\label{eqn:mm-adagrad:bound}
\nonumber  &\quad {\bf Reg}_T^d(\bbx^*_1,\ldots,\bbx^*_T)
   \\&=\min\left\{
   \begin{array}{ll}
      \varpi_1 \sum_{i=1}^p \|\tilde{g}_{1:T,i} \|
 +\varpi_2  \sum_{i=1}^{p}(\tilde{\upsilon}_{T,i})^{1/2} {C}^*_{T,i}+\varpi_3, \\
 \\
     \frac{1}{2\gamma}\sum_{t=1}^{T}\| \nabla f_t(\bbx_t^*)\|^2+\acute{\varpi}_1 \sum_{i=1}^p \|\tilde{g}_{1:T,i} \|
 +\acute{\varpi}_2   \sum_{i=1}^{p}(\tilde{\upsilon}_{T,i})^{1/2}{S}^*_{T,i}+\acute{\varpi}_3,
   \end{array}
 \right.
\end{align}
where $\bbx_t^* =\argmin_{\bbx \in \mathcal{X}} f_t(\bbx)$, ${C}^*_{T,i}$ and ${S}^*_{T,i}$ are defined as in \eqref{CTT} and \eqref{STT}, respectively, and
\begin{align}\label{sgmm}
\nonumber& \varpi_1:=1+\frac{4\tilde{\sigma} \beta \alpha \theta}{3(1-\beta)^3},\quad  \varpi_2:=\frac{2}{3}\big(\frac{\tilde{\sigma}\beta^2 \theta }{1-\beta}+2D_{\infty} \big),\quad \tilde{\sigma}:=\frac{2}{\lambda+\frac{1}{\alpha}},\quad \theta:=\frac{1}{2\lambda \alpha}+\frac{1}{2},
\\\nonumber& \varpi_3:=\frac{2}{3}\Big(\sum_{i=1}^p (\tilde{\upsilon}_{1,i})^{1/2}(x_{1,i}- x_{1,i}^*)^2
+\big( \frac{\tilde{\sigma}\beta\theta }{2\alpha(1-\beta)}+1\big)D_{\infty}^2 \sum_{i=1}^{p}(\tilde{\upsilon}_{T,i})^{1/2}\Big),
\\\nonumber& \acute{\varpi}_1:={\tilde{\sigma}\theta}(L+\gamma)\frac{4\beta\alpha}{(1 - \beta)^3},\quad \acute{\varpi}_2:=(L+\gamma)\big(\frac{\beta\tilde{\sigma}\theta }{1-\beta}+2\big) ,\\
& \acute{\varpi}_3:=(L+\gamma)\Big(\sum_{i=1}^p (\tilde{\upsilon}_{1,i})^{1/2}(x_{1,i}- x_{1,i}^*)^2+
\big(\frac{\beta\tilde{\sigma} \theta }{\alpha(1-\beta)}
+1 \big)D_{\infty}^2\sum_{i=1}^{p}(\tilde{\upsilon}_{T,i})^{1/2}\Big).
\end{align}
Here, $(\tilde{\upsilon}_{T,i})^{1/2}:=\max_{0\leq l\leq K-1}({\upsilon}_{T,i}^{K-l})^{1/2}$ and $\|\tilde{g}_{1:T,i} \|:=\max_{0\leq l \leq K-1} \|g^{K-l}_{1:T,i} \|$.
\end{thm}
\begin{proof}
By the $L$-smoothness of $f_t(\cdot)$ and the fact that $2\langle a, b\rangle\leq \gamma\| a\|^2+\gamma^{-1}\| b\|^2$ for any $\gamma>0$, we have
  \begin{align*}
 \nonumber f_t(\bbx_t) -f_t(\bbx_t^*)  & \leq \langle \nabla f_t(\bbx_t^*),\bbx_t-\bbx_t^* \rangle+\frac{L}{2}\|\bbx_t-\bbx_t^* \|^2_{({\bbv}_t^{K})^{1/2}} \\
 \nonumber    & \leq \frac{1}{2\gamma}\| \nabla f_t(\bbx_t^*)\|^2_{({\bbv}_t^{K})^{-1/2}}+\frac{\gamma}{2} \|\bbx_t-\bbx_t^* \|^2_{({\bbv}_t^{K})^{1/2}}+\frac{L}{2}\|\bbx_t-\bbx_t^* \|^2_{({\bbv}_t^{K})^{1/2}}\\
     & =\frac{1}{2\gamma}\| \nabla f_t(\bbx_t^*)\|^2_{({\bbv}_t^{K})^{-1/2}}+\frac{(L+\gamma)}{2} \|\bbx_t-\bbx_t^* \|^2_{({\bbv}_t^{K})^{1/2}}.
  \end{align*}
Summing from $t = 1$ to $t = T$ on both sides yields:
\begin{align}\label{ew}
  \sum_{t=1}^{T}\big( f_t(\bbx_t) -f_t(\bbx_t^*)\big) &\leq \frac{1}{2\gamma}\sum_{t=1}^{T}\| \nabla f_t(\bbx_t^*)\|^2_{({\bbv}_t^{K})^{-1/2}}+\frac{(L+\gamma)}{2} \sum_{t=1}^{T}\|\bbx_t-\bbx_t^* \|^2_{({\bbv}_t^{K})^{1/2}}.
\end{align}
We next bound term $\sum_{t=1}^{T}\|\bbx_t-\bbx_t^* \|^2_{({\bbv}_t^{K})^{1/2}}$.
Note that
\begin{align}\label{weq}
\nonumber &\quad\sum_{t=1}^{T}\|\bbx_t-\bbx_t^* \|^2_{({\bbv}_t^{K})^{1/2}}
\\\nonumber &=\sum_{t=1}^{T}\big(\|\bbx_t-\bbx_t^* \|^2_{({\bbv}_t^{K})^{1/2}}
-\|\bbx_{t+1}-\bbx_{t+1}^* \|^2_{({\bbv}_t^{K})^{1/2}}\big)+\sum_{t=1}^{T}\|\bbx_{t+1}-\bbx_{t+1}^* \|^2_{({\bbv}_t^{K})^{1/2}}
\\\nonumber&\leq \|\bbx_1-\bbx_1^* \|^2_{({\bbv}_1^{K})^{1/2}}
+ D_{\infty}^2\sum_{i=1}^p  ({\upsilon}_{T,i}^K)^{1/2}
+2\sum_{t=1}^{T}\|\bbx_{t+1}-\bbx_{t}^* \|^2_{({\bbv}_t^{K})^{1/2}}+2\sum_{t=1}^{T}\|\bbx_{t}^*-\bbx_{t+1}^* \|^2_{({\bbv}_t^{K})^{1/2}}
\\&= \|\bbx_1-\bbx_1^* \|^2_{({\bbv}_1^{K})^{1/2}}
+ D_{\infty}^2\sum_{i=1}^p  ({\upsilon}_{T,i}^K)^{1/2}
+2\sum_{t=1}^{T}\|\bbx_{t+1}-\bbx_{t}^* \|^2_{({\bbv}_t^{K})^{1/2}}+2\sum_{i=1}^{p}(\upsilon_{T,i}^K)^{1/2}{S}^*_{T,i},
\end{align}
where the inequality follows from \eqref{weq1} and $\|a+b\|^2\leq 2\| a\|^2+ 2\|b\|^2$, the last equality is by Eq. \eqref{STT}.
\\
Recall the Lemma \ref{lem:strong:convex2}:
\begin{align*}
\nonumber & \quad  \sum_{t=1}^{T} \sum_{i=1}^p {({\upsilon}^{j}_t)^{1/2}}(z_{t,i}^{j+1}- x_{t,i}^*)^2
\\&\leq {\bar{\sigma}}\sum_{t=1}^{T} \sum_{i=1}^p  {({\upsilon}^{j}_t)^{1/2}}(z_{t,i}^{j}- x_{t,i}^*)^2
\\&+\frac{\beta \tilde{\sigma}}{1-\beta}\Big(\sum_{i=1}^{p}\big(\frac{ D_{\infty}^2}{2\alpha}+\frac{S^*_{T,i}}{2}\big) ({\upsilon}_{T,i}^j)^{1/2}
+\frac{2\alpha}{(1 - \beta)^2} \sum_{i=1}^p \|g^j_{1:T,i} \|\Big),
\end{align*}
which implies
\begin{align*}
\nonumber &\quad \sum_{t=1}^{T}\|\bbx_{t+1}-\bbx_{t}^* \|^2_{({\bbv}_t^{K})^{1/2}} =\sum_{t=1}^{T}\|\bbz_{t}^{K+1}-\bbx_{t}^* \|^2_{({\bbv}_t^{K})^{1/2}}
\\\nonumber&\leq {\bar{\sigma}}^K\sum_{t=1}^{T}\|\bbx_{t}-\bbx_{t}^*\|^2_{({\bbv}_t^{K})^{1/2}}+\frac{\beta \tilde{\sigma}}{1-\beta}
\sum_{i=1}^{p}\big(\frac{ D_{\infty}^2}{2\alpha}+\frac{S^*_{T,i}}{2}\big) \sum_{l=0}^{K-1}{\bar{\sigma}}^{l}({\upsilon}_{T,i}^{K-l})^{1/2}
\\&+\frac{2\beta\alpha}{(1 - \beta)^3}\tilde{\sigma} \sum_{l=0}^{K-1}{\bar{\sigma}}^{l}\sum_{i=1}^p \|g^{K-l}_{1:T,i} \|
\\\nonumber&\leq {\bar{\sigma}}^K\sum_{t=1}^{T}\|\bbx_{t}-\bbx_{t}^*\|^2_{({\bbv}_t^{K})^{1/2}}+\frac{\beta \tilde{\sigma}}{1-\beta}
\sum_{i=1}^{p} \big(\frac{ D_{\infty}^2}{2\alpha}+\frac{S^*_{T,i}}{2}\big) \max_{0\leq l\leq K-1}({\upsilon}_{T,i}^{K-l})^{1/2}\sum_{l=0}^{K-1}{\bar{\sigma}}^{l}
\\&+\frac{2\beta\alpha}{(1 - \beta)^3}\tilde{\sigma} \max_{0\leq l\leq K-1}\sum_{i=1}^p \|g^{K-l}_{1:T,i} \|\sum_{l=0}^{K-1}{\bar{\sigma}}^{l}.
\end{align*}
Taking $K=\lceil \frac{1/{\alpha}+\lambda}{2\lambda}\ln 4\rceil$ such that
\begin{subequations}
\begin{align}
\label{19}{\bar{\sigma}}^K&= \big(1-\frac{2\lambda}{\frac{1}{\alpha}+\lambda}\big)^K\leq \exp (-\frac{2K\lambda}{\frac{1}{\alpha}+\lambda})\leq \frac{1}{4},\\
\label{20}\sum_{l=0}^{K-1}{\bar{\sigma}}^l&=\sum_{l=0}^{K-1} \big(1-\frac{2\lambda}{\frac{1}{\alpha}+\lambda}\big)^l\leq \frac{1}{2\lambda \alpha}+\frac{1}{2}:=\theta,
\end{align}
\end{subequations}
the inequality yields
\begin{align}\label{wew}
\sum_{t=1}^{T}\|\bbx_{t+1}-\bbx_{t}^* \|^2_{({\bbv}_t^{K})^{1/2}} &  \leq \frac{1}{4}\sum_{t=1}^{T}\|\bbx_{t}-\bbx_{t}^*\|^2_{({\bbv}_t^{K})^{1/2}}
  +\frac{\beta \tilde{\sigma}\theta\varphi}{1-\beta},
\end{align}
where
$$\varphi:= \sum_{i=1}^{p}\big(\frac{ D_{\infty}^2}{2\alpha}+\frac{{S}^*_{T,i}}{2}\big)\max_{0\leq l\leq K-1}({\upsilon}_{T,i}^{K-l})^{1/2}+\frac{2\alpha}{(1 - \beta)^2} \max_{0\leq l\leq K-1}\sum_{i=1}^p \|g^{K-l}_{1:T,i} \|.$$
Injecting \eqref{wew} into \eqref{weq} gives us
\begin{align*}
  \sum_{t=1}^{T}\|\bbx_t-\bbx_t^*\|^2_{({\bbv}_t^{K})^{1/2}} &\leq \|\bbx_1-\bbx_1^*\|^2_{({\bbv}_1^{K})^{1/2}}+ D_{\infty}^2\sum_{i=1}^{p}(\upsilon_{T,i}^K)^{1/2}+\frac{1}{2}\sum_{t=1}^{T}\|\bbx_{t}-\bbx_{t}^*\|^2_{({\bbv}_t^{K})^{1/2}}
  \\&+\frac{2\beta \tilde{\sigma}\theta\varphi}{1-\beta}+2\sum_{i=1}^{p}(\upsilon_{T,i}^K)^{1/2}{S}^*_{T,i} .
\end{align*}
Rearranging the above terms, we have
\begin{align*}
   \sum_{t=1}^{T}\|\bbx_t-\bbx_t^*\|^2_{({\bbv}_t^{j})^{1/2}} & \leq 2\|\bbx_1-\bbx_1^*\|^2_{({\bbv}_1^{j})^{1/2}}+\frac{4\beta \tilde{\sigma}\theta\varphi}{1-\beta}+2 D_{\infty}^2\sum_{i=1}^{p}(\upsilon_{T,i}^j)^{1/2}+4\sum_{i=1}^{p}(\upsilon_{T,i}^j)^{1/2}{S}^*_{T,i}.
\end{align*}
Plugging the above inequality into \eqref{ew}, we get, for all $\gamma\geq 0$,
\begin{align*}
  \sum_{t=1}^{T} \big(f_t(\bbx_t) -f_t(\bbx_t^*) \big) & \leq \frac{1}{2\gamma}\sum_{t=1}^{T}\| \nabla f_t(\bbx_t^*)\|^2+(L+\gamma)\|\bbx_1-\bbx_1^*\|^2_{({\bbv}_1^{K})^{1/2}} \\& +(L+\gamma)\frac{2\beta \tilde{\sigma}\theta\varphi}{1-\beta}
  +(L+\gamma)\sum_{i=1}^{p}\big(D_{\infty}^2+2{S}^*_{T,i} \big) (\upsilon_{T,i}^K)^{1/2}.
\end{align*}

In addition, we demonstrate that the dynamic regret is still upper-bounded by ${C}^*_{T,i}$.
Note that from the update rule of Algorithm~\ref{alg:subgradient2} we have
$$\bbz_{t}^{j+1}=\Pi_{\mathcal{X}, ({\bbv}^j_t)^{1/2}}\big(\bbz_{t}^j - \alpha({\bbv}^j_t)^{-1/2} \bbm^j_t\big),\quad j=1,\ldots,K.$$
By the same argument leading to \eqref{eqn:thm1:11}, we have
\begin{align} \label{eqn:thm1:11d}
\nonumber {\bf Reg}_T^d(\bbx^*_1,\ldots,\bbx^*_T)&=\sum_{t=1}^T \big(f_{t}(\bbx_{t}) - f_{t}(\bbx_{t}^*)\big)
\\&\leq \max_{0\leq l \leq K-1}\sum_{i=1}^p \|g^{K-l}_{1:T,i} \|+\frac{1}{2}\sum_{t=1}^T\|\bbx_{t} - \bbx_{t}^*\|_{(\bbv_t^K)^{1/2}}^2.
\end{align}
For the second term on the RHS of \eqref{eqn:thm1:11d}, we have
\begin{align}\label{nkj2}
\nonumber  &\quad\sum_{t=1}^T \sum_{i=1}^p  (\upsilon_{t,i}^K)^{1/2}(x_{t,i} - x_{t,i}^*)^2
\\\nonumber&=\sum_{t=1}^T \sum_{i=1}^p\big( (\upsilon_{t,i}^K)^{1/2}(x_{t+1,i} - x_{t+1,i}^*)^2-(\upsilon_{t,i}^K)^{1/2}(x_{t+1,i} - x_{t,i}^*)^2\big)
\\\nonumber&+\sum_{t=1}^T \sum_{i=1}^p\big( (\upsilon_{t,i}^K)^{1/2}(x_{t,i} - x_{t,i}^*)^2-(\upsilon_{t,i}^K)^{1/2}(x_{t+1,i} - x_{t+1,i}^*)^2\big)
\\
&+\sum_{t=1}^T \sum_{i=1}^p (\upsilon_{t,i}^K)^{1/2}(x_{t+1,i} - x_{t,i}^*)^2 .
\end{align}
We first setup an upper bound for the last term on the RHS of \eqref{nkj2}.
To this end, note that applying the Lemma \ref{lem:strong:convex} yields:
\begin{align}\label{asdas}
 \sum_{t=1}^{T} \sum_{i=1}^p ({\upsilon}^{j}_{t,i})^{1/2}(z^{j+1}_{t,i}- x_{t,i}^*)^2
&\leq
\bar{\sigma}\sum_{t=1}^{T} \sum_{i=1}^p ({\upsilon}^j_{t,i})^{1/2}(z^j_{t,i}- x_{t,i}^*)^2
+\frac{\beta \tilde{\sigma}\vartheta}{1-\beta},
\end{align}
where
$$\vartheta:=\sum_{i=1}^{p}\big(\frac{ D_{\infty}^2}{2\alpha}+\beta {C}^*_{T,i}\big) ({\upsilon}_{T,i}^j)^{1/2}+\frac{2\alpha}{(1 - \beta)^2} \sum_{i=1}^p \|g_{1:T,i}^j \|.$$
Thus, in light of Eq. \eqref{asdas}, we obtain that
\begin{align*}
\nonumber &\quad \sum_{t=1}^{T}\|\bbx_{t+1}-\bbx_{t}^* \|^2_{({\bbv}_t^{K})^{1/2}} =\sum_{t=1}^{T}\|\bbz_{t}^{K+1}-\bbx_{t}^* \|^2_{({\bbv}_t^{K})^{1/2}}
\\&\leq {\bar{\sigma}}^K\sum_{t=1}^{T}\|\bbx_{t}-\bbx_{t}^*\|^2_{({\bbv}_t^{K})^{1/2}}
+\frac{\beta \tilde{\sigma}}{1-\beta}
\sum_{i=1}^{p}\big(\frac{ D_{\infty}^2}{2\alpha}+\beta {C}^*_{T,i}\big)\sum_{l=0}^{K-1}{\bar{\sigma}}^{l} ({\upsilon}_{T,i}^{K-l})^{1/2}
\\&+\frac{2\beta \tilde{\sigma} \alpha}{(1 - \beta)^3} \sum_{l=0}^{K-1}{\bar{\sigma}}^{l}\sum_{i=1}^p \|g^{K-l}_{1:T,i} \|
\\\nonumber&\leq {\bar{\sigma}}^K\sum_{t=1}^{T}\|\bbx_{t}-\bbx_{t}^*\|^2_{({\bbv}_t^{K})^{1/2}}
+\frac{\beta \tilde{\sigma}}{1-\beta} \sum_{i=1}^{p}\big(\frac{ D_{\infty}^2}{2\alpha}+\beta {C}^*_{T,i}\big) \max_{0\leq l\leq K-1}({\upsilon}_{T,i}^{K-l})^{1/2}\sum_{l=0}^{K-1}{\bar{\sigma}}^{l}
\\&+\frac{2\beta \tilde{\sigma} \alpha}{(1 - \beta)^3}\max_{0\leq l \leq K-1}\sum_{i=1}^p \|g^{K-l}_{1:T,i} \| \sum_{l=0}^{K-1}{\bar{\sigma}}^{l}.
\end{align*}
Next, according to \eqref{19} and \eqref{20}, we get
\begin{align}\label{bghh}
\sum_{t=1}^{T}\|\bbx_{t+1}-\bbx_{t}^* \|^2_{({\bbv}_t^{K})^{1/2}}&\leq \frac{1}{4}\sum_{t=1}^{T}\|\bbx_{t}-\bbx_{t}^*\|^2_{({\bbv}_t^{K})^{1/2}}
 +\frac{\beta \tilde{\sigma}\hat{\vartheta}}{1-\beta},
\end{align}
where
$$\hat{\vartheta}=\theta\Big(\sum_{i=1}^{p}\big(\frac{ D_{\infty}^2}{2\alpha}+\beta {C}^*_{T,i}\big)\max_{0\leq l\leq K-1}({\upsilon}_{T,i}^{K-l})^{1/2}+\frac{2\alpha }{(1 - \beta)^2} \max_{0\leq l \leq K-1}\sum_{i=1}^p \|g^{K-l}_{1:T,i} \|\Big).$$
Then, substituting \eqref{bghh}, \eqref{re}, and \eqref{weq1} into \eqref{nkj2}, we get
\begin{align*}
\nonumber \sum_{t=1}^{T}\|\bbx_{t}-\bbx_{t}^*\|^2_{({\bbv}_t^{K})^{1/2}} &\leq
\sum_{i=1}^p ({\upsilon}_{1,i}^K)^{1/2}(x_{1,i}- x_{1,i}^*)^2
 +D_{\infty}^2 \sum_{i=1}^{p}(\upsilon_{T,i}^K)^{1/2}
+2D_{\infty}\sum_{i=1}^{p} ({\upsilon}_{T,i}^K)^{1/2}{C}^*_{T,i}
\\&+\frac{1}{4}\sum_{t=1}^{T}\|\bbx_{t}-\bbx_{t}^*\|^2_{({\bbv}_t^{K})^{1/2}}
 +\frac{\beta \tilde{\sigma}\hat{\vartheta}}{1-\beta}.
\end{align*}
Rearranging the terms, we get
\begin{align*}
\nonumber \sum_{t=1}^{T}\|\bbx_{t}-\bbx_{t}^*\|^2_{({\bbv}_t^{K})^{1/2}} &\leq
\frac{4}{3}\Big(\sum_{i=1}^p ({\upsilon}_{1,i}^K)^{1/2}(x_{1,i}- x_{1,i}^*)^2
 +D_{\infty}^2 \sum_{i=1}^{p}(\upsilon_{T,i}^K)^{1/2}
\\&+2D_{\infty}\sum_{i=1}^{p} ({\upsilon}_{T,i}^K)^{1/2}{C}^*_{T,i}
 +\frac{\beta \tilde{\sigma}\hat{\vartheta}}{1-\beta}\Big).
\end{align*}
Combining the above inequality and \eqref{eqn:thm1:11d}, we observe that
\begin{align*}
\nonumber\sum_{t=1}^T \big(f_{t}(\bbx_{t}) - f_{t}(\bbx_{t}^*)\big)
&\leq \max_{0\leq l \leq K-1}\sum_{i=1}^p \|g^{K-l}_{1:T,i} \|
+\frac{2}{3}\Big(\sum_{i=1}^p ({\upsilon}_{1,i}^K)^{1/2}(x_{1,i}- x_{1,i}^*)^2
\\&
+D_{\infty}^2 \sum_{i=1}^{p}(\upsilon_{T,i}^K)^{1/2}
+2D_{\infty}\sum_{i=1}^{p} ({\upsilon}_{T,i}^K)^{1/2}{C}^*_{T,i} +\frac{\beta \tilde{\sigma}\hat{\vartheta}}{1-\beta}\Big).
\end{align*}
This completes the proof.
\end{proof}
\section{Experiments} \label{exper}
To validate our new theoretical results, we have conducted a simple experiment on the online quadratic optimization with a sequence of optimal values.
\begin{figure}[ht]
\includegraphics[width=9cm]{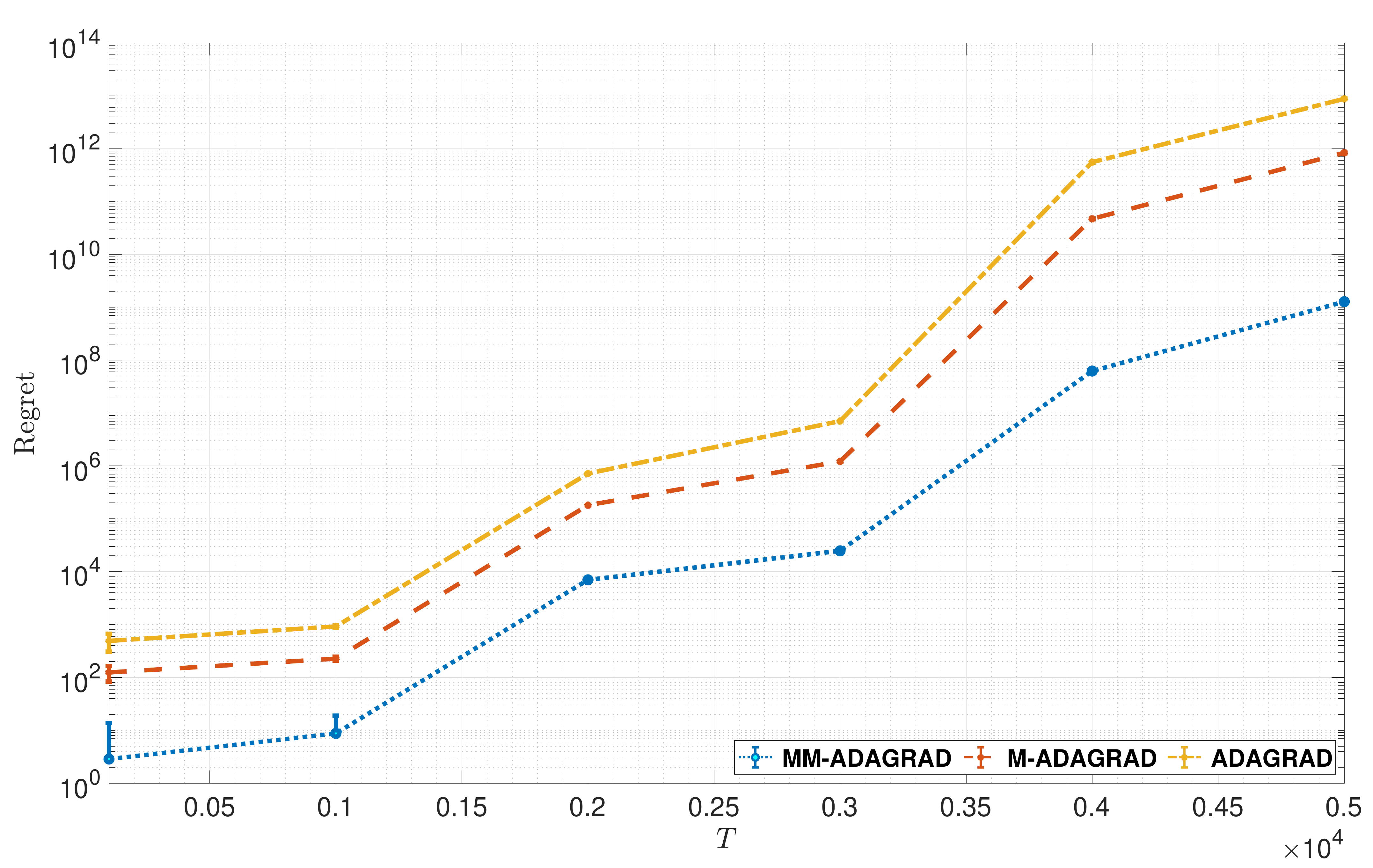}
\centering
        \caption{ Performance of  \textsc{AdaGrad}-Type methods  with three different minimizers $\{\mathbf{x}_t^*\}_{t=1}^3$. }
        \label{fig:dynamic}
\end{figure}
\begin{figure}[ht]
\includegraphics[width=9cm]{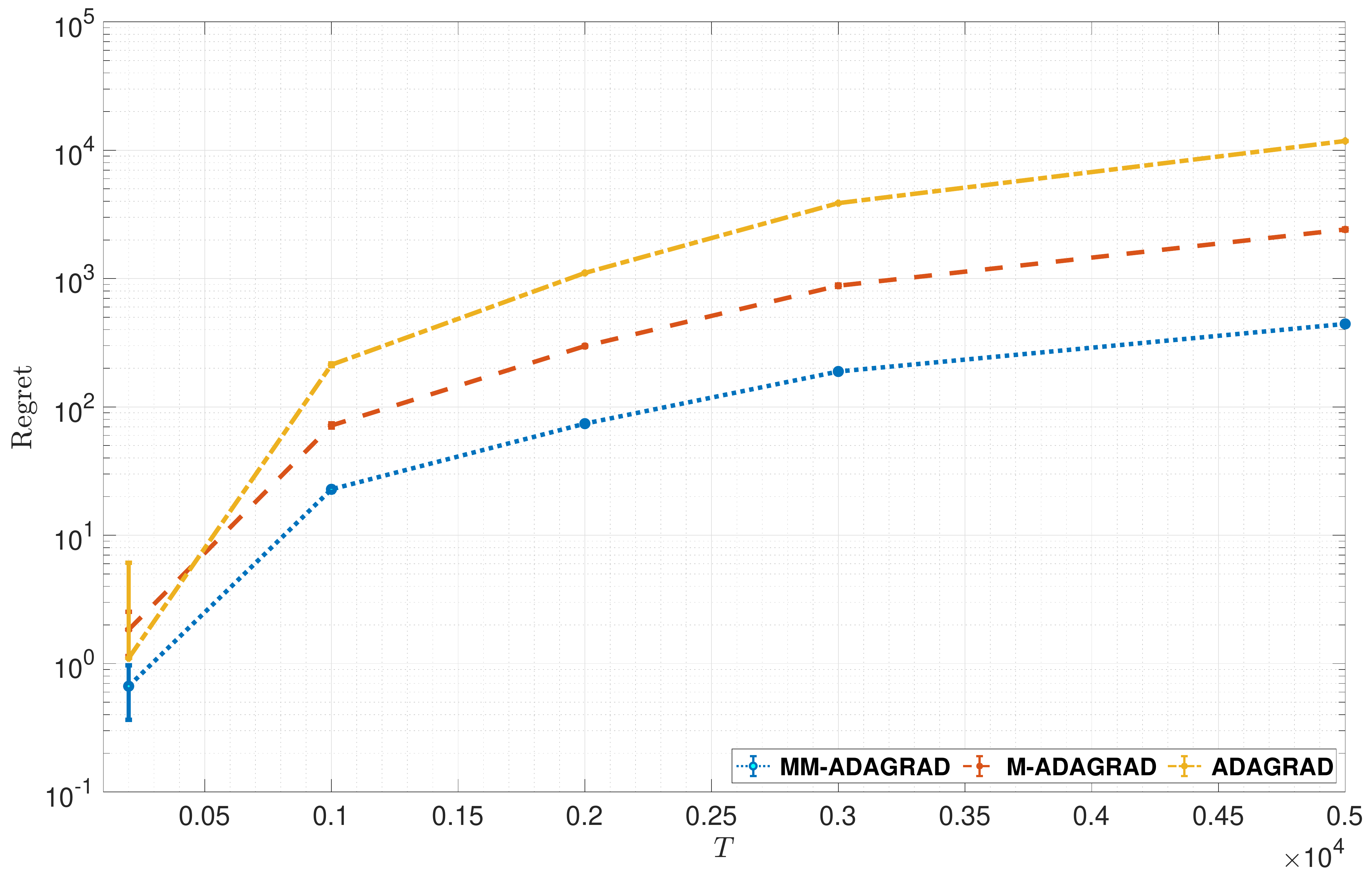}
\centering
        \caption{ Performance of \textsc{AdaGrad}-Type methods on online static regression ($\mathbf{x}= \mathbf{x}^*_t$ for all $t \in [T]$). }
        \label{fig:static}
\end{figure}

We have simulated the online learning scenario by the following setting: the player sequentially receives the feature of data item and then predict its label. The data item of each round is denoted by $(\mathbf{a}_t, b_t) \in \mathbb{R}^{p} \times \mathbb{R}$. The time horizon is set to $T=5000$. To simulate the distribution changes, we generate the output according to $b_t = \mathbf{a}_t^\top \mathbf{x}_t^*+ \epsilon_t$, where $\mathbf{x}_t^*\in \mathbb{R}^{p}$ is the underlying model and $\epsilon_t \in [0,0.1]$ is the random noise. The underlying model $\mathbf{x}_t^*$ will change every $2000$ rounds, randomly sampled from a $p$-dimensional ball with diameter $D_{\infty}=5$, so there are in total three changes. We have chosen the loss function as the square loss, defined as $f_t(\mathbf{x})=\frac{1}{2} (\mathbf{a}^\top_t \mathbf{x}-b_t)^2$. We set $K=10$, $\beta=0.9$, and $\alpha=0.001$.

Figure~\ref{fig:dynamic} demonstrates the performance of  $\textsc{AdaGrad}$, $\textsc{M-AdaGrad}$, and $\textsc{MM-AdaGrad}$. It can be easily seen that using multiple gradient updates at each time step $t$ ($\textsc{MM-AdaGrad}$) improves the regret bound.  Further,  using momentum update ($\textsc{M-AdaGrad}$) can accelerate the performance of $\textsc{AdaGrad}$ in the dynamic environment. 

Figure~\ref{fig:static} also illustrates the performance of \textsc{MM-AdaGrad} in the static setting, that is $\mathbf{x}= \mathbf{x}^*_t$ for all $t \in [T]$. Similar to the dynamic setting,  \textsc{MM-AdaGrad} achieves the best regret bound while \textsc{AdaGrad} gives the worst performance.
\section{Conclusion}\label{conclusion}

In this paper, we have investigated the dynamic regret of a variant of \textsc{AdaGrad} adapted to strongly convex and smooth functions.
We first have proposed Momentum \textsc{AdaGrad} (M-\textsc{AdaGrad}), which achieves a regret bound of $O(C^*_T)$ with one gradient query per round.
Next, we have developed the Multiple Momentum \textsc{AdaGrad} (MM-\textsc{AdaGrad}) method, which achieves a regret bound of $O(\min\{C^*_T,S^*_T\})$ under some mild sufficient conditions with multiple gradient queries per round.
Numerical results reveal the efficiency and effectiveness of the proposed methods in practice.

\newpage
\begin{center}
{\Large
\textsc{Appendix}
}
\end{center}
In this section we provide lemmas that will be used to prove our main theorem.
\subsection*{Details of Section \ref{sec:main}}

\begin{lem}\cite{hazan2011beyond} \label{lem:support} Assume that $f: \mathcal{X} \mapsto \mathbb{R}$ is $\lambda$-strongly convex, and $\bbx^* =\argmin_{\bbx \in \mathcal{X}} f(\bbx)$. Then, we have
\begin{equation*}
f(\bbx)-f(\bbx^*) \geq \frac{\lambda}{2} \|\bbx-\bbx^*\|^2, \ \forall \bbx \in \mathcal{X}.
\end{equation*}
\end{lem}

\begin{lem}\cite[Lemma 5]{mcmahan2010adaptive}
\label{mc}
 For any $\bbq\in \mathbb{R}^p$ and convex feasible set $\mathcal{X}\subset \mathbb{R}^p$, let
\begin{eqnarray*}
\bbu_1 &\leftarrow& \argmin_{\bbx\in \mathcal{X}}\|\bbx-\bbz_1\|_{\bbq},\\
\bbu_2 &\leftarrow& \argmin_{\bbx\in \mathcal{X}}\|\bbx-\bbz_2\|_{\bbq}.
\end{eqnarray*}
Then, we have $$\|\bbu_1-\bbu_2\|_{\bbq}\leq \| \bbz_1-\bbz_2\|_{\bbq}.$$
\end{lem}

\begin{lem}\cite[Lemma 3.5]{auer2002adaptive}
\label{lem:simple-grad-boundd}
For any non-negative real numbers $\bby_1, \ldots, \bby_t$, the following holds:
$$
\sum_{r=1}^t \frac{\bby_r}{\sqrt{\sum_{s=1}^r \bby_s}} \leq 2 \sqrt{\sum_{r=1}^t \bby_r}.
$$
\end{lem}

\begin{lem}\cite[Lemma 1]{alacaoglu2020new}
\label{lem:simple-grad-bound}
Let $\bbm_{t}= \beta \bbm_{t-1}+(1-\beta)\bbg_{t}$. Then, for any vectors $A_{t-1}$ and $A_t$, we have
\begin{equation*}
 \langle A_t,\bbg_t \rangle=\frac{1}{1-\beta}(\langle A_t,\bbm_{t} \rangle-\langle A_{t-1},\bbm_{t-1} \rangle)+\langle A_{t-1},\bbm_{t-1} \rangle+\frac{\beta}{1-\beta}\langle A_{t-1}-A_t,\bbm_{t-1} \rangle.
\end{equation*}
\end{lem}
\begin{lem}
\label{lem: first_lem_sadam}
For $\bbm_{t}$ and ${\bbv}_{t}$ generated by Algorithm~\ref{alg:subgradient}, we have
\begin{equation*}
\sum_{t=1}^{T} \| \bbm_{t}\|_{\bbv_t^{-1/2}} ^ 2 \leq \frac{2}{(1 - \beta)^2} \sum_{i=1}^p \|g_{1:T,i} \|.
\end{equation*}
\end{lem}

\begin{proof}
The proof is similar to that of \cite[Lemma 3]{reddi2018convergence}.
Observe that:
\begin{align*}\label{fd}
\nonumber\sum_{t=1}^{T}  \| \bbm_{t}\|_{\bbv_t^{-1/2}} ^ 2 &= \sum_{t=1}^{T-1}  \| \bbm_{t}\|_{\bbv_t^{-1/2}} ^ 2 +  \sum_{i=1}^p \frac{m_{T,i} ^ 2}{\sqrt{\upsilon_{T,i}}} \\\nonumber
&\stackrel{(i)}= \sum_{t=1}^{T-1}  \| \bbm_{t}\|_{\bbv_t^{-1/2}} ^ 2 +   \sum_{i=1}^p \frac{\big(  \sum_{j=1}^T  (1 - \beta_{j})\Pi_{k=1}^{T-j}\beta_{(T-k+1)} g_{j,i}\big) ^ 2}{\sqrt{ \sum_{j=1}^T g_{j,i}^2}}\\
&\stackrel{(ii)}\leq \sum_{t=1}^{T-1}  \| \bbm_{t}\|_{\bbv_t^{-1/2}}^ 2 +   \sum_{i=1}^p \frac{\big(  \sum_{j=1}^T  \Pi_{k=1}^{T-j}\beta_{(T-k+1)}\big)
\big(  \sum_{j=1}^T  \Pi_{k=1}^{T-j}\beta_{(T-k+1)}g_{j,i}^2\big)}{\sqrt{ \sum_{j=1}^T g_{j,i}^2}}\nonumber \\
&\stackrel{(iii)}\leq \sum_{t=1}^{T-1}  \|\bbm_{t}\|_{\bbv_t^{-1/2}} ^ 2 +  \sum_{i=1}^p \frac{(\sum_{j=1}^{T}\beta^{T-j}) (\sum_{j=1}^T \beta^{T - j} g_{j,i}^2)}{\sqrt{ \sum_{j=1}^T g_{j,i}^2}}\nonumber \\
&\stackrel{(iv)}\leq \sum_{t=1}^{T-1}  \|\bbm_{t}\|_{\bbv_t^{-1/2}} ^ 2 +  \frac{1}{1-\beta} \sum_{i=1}^p \frac{ \sum_{j=1}^T \beta^{T - j} g_{j,i}^2}{\sqrt{ \sum_{j=1}^T g_{j,i}^2}}\nonumber \\
&\leq \sum_{t=1}^{T-1}  \| \bbm_{t}\|_{\bbv_t^{-1/2}} ^ 2 +  \frac{1}{(1-\beta)} \sum_{i=1}^p \sum_{j=1}^T\frac{ \beta^{T - j} g_{j,i}^2}{\sqrt{ \sum_{k=1}^j  g_{k,i}^2}},
\end{align*}
where (i) follows from the update rule of Algorithm~\ref{alg:subgradient}; (ii) follows from Cauchy-Schwarz inequality; (iii) follows from the inequality  $\beta_{k}\leq \beta$ for all $k\in [T]$; (iv) follows from $\sum_{j=1}^{T}\beta^{T-j}\leq 1/(1-\beta)$.
As a result,
\begin{align*}\sum_{t=1}^{T}  \| \bbm_{t}\|_{\bbv_t^{-1/2}} ^ 2
&\leq   \frac{1}{(1-\beta)} \sum_{i=1}^p \sum_{j=1}^T\frac{\sum_{l=0}^{T-j} \beta^{l} g_{j,i}^2}{\sqrt{ \sum_{k=1}^j  g_{k,i}^2}}\nonumber \\
&\leq   \frac{1}{(1-\beta)^2} \sum_{i=1}^p \sum_{j=1}^T\frac{ g_{j,i}^2}{\sqrt{ \sum_{k=1}^j  g_{k,i}^2}}\nonumber \\
&\leq   \frac{2}{(1-\beta)^2} \sum_{i=1}^p \|g_{1:T,i} \|,
\end{align*}
where the last inequality holds due to Lemma~\ref{lem:simple-grad-boundd}.
This completes the proof of the lemma.
\end{proof}
\begin{lem} \label{lem:strong:convexsss}
The decision sequence $\{\bbx_{t}\}$ generated by Algorithm~\ref{alg:subgradient} satisfies
\begin{align*}
 \langle \bbm_t,\bbx_{t+1}-\bbx_t^*\rangle &\leq \frac{1}{2\alpha}\|\bbx_t^*-\bbx_t\|_{{\bbv}^{1/2}_{t}}^2 -\frac{1}{2\alpha}\|\bbx_t^*-\bbx_{t+1}\|_{{\bbv}^{1/2}_{t}}^2-\frac{1}{2\alpha}\|\bbx_{t+1}-\bbx_t\|_{{\bbv}^{1/2}_{t}}^2,
\end{align*}
where $\bbx_t^* =\argmin_{\bbx \in \mathcal{X}} f_t(\bbx).$
\end{lem}
\begin{proof}
The update rule of $\bbx_{t+1}$ in Algorithm~\ref{alg:subgradient} can be rewritten equivalently as
\begin{equation}\label{eee}
\bbx_{t+1}= \argmin_{\bbx \in \mathcal{X} } \Big\{\ f_t(\bbx_t) + \langle  \bbm_t, \bbx- \bbx_t \rangle  + \frac{1}{2 \alpha} \|\bbx- \bbx_t\|_{{\bbv}^{1/2}_{t}}^2\Big\}.
\end{equation}
From the optimality condition of \eqref{eee}, we have
$$0\in \mathcal{N}_{\mathcal{X}}(\bbx_{t+1})+{\bbv}^{1/2}_{t} (\bbx_{t+1}-\bbx_{t})+\alpha \bbm_t ,$$
where $\mathcal{N}_{\mathcal{X}}$ denotes the normal cone of $\mathcal{X}$ at $x$. Hence, it follows
\begin{equation}\label{dd}
\langle \bbx_{t+1}-\bbx,{\bbv}^{1/2}_{t} (\bbx_{t+1}-\bbx_{t})+\alpha \bbm_t \rangle \leq 0,\,\, \forall \bbx\in \mathcal{X}.
\end{equation}
In addition, it holds
$$ \langle \bbx_{t+1}-\bbx,{\bbv}^{1/2}_{t} (\bbx_{t+1}-\bbx_{t}) \rangle=\frac{1}{2}
(  \|\bbx_{t+1}- \bbx_t\|_{{\bbv}^{1/2}_{t}}^2
-  \|\bbx- \bbx_t\|_{{\bbv}^{1/2}_{t}}^2
+  \|\bbx_{t+1}- \bbx\|_{{\bbv}^{1/2}_{t}}^2).$$
Substituting the above  equation into \eqref{dd} and setting $\bbx=\bbx_t^*$, gives
\begin{align*}
 \langle \bbm_t,\bbx_{t+1}-\bbx_t^*\rangle &\leq \frac{1}{2\alpha}\|\bbx_t^*-\bbx_t\|_{{\bbv}^{1/2}_{t}}^2 -\frac{1}{2\alpha}\|\bbx_t^*-\bbx_{t+1}\|_{{\bbv}^{1/2}_{t}}^2-\frac{1}{2\alpha}\|\bbx_{t+1}-\bbx_t\|_{{\bbv}^{1/2}_{t}}^2.
\end{align*}
This completes the proof.
\end{proof}
\subsubsection*{Proof of Lemma \ref{lem:strong:convex}}
\begin{proof}
By the strong convexity of function $f_t$ it holds that
\begin{align}\label{sdc}
\nonumber \sum_{t=1}^T \big(f_{t}(\bbx_{t}) - f_{t}(\bbx_{t}^*)\big)
&\leq \sum_{t=1}^T\langle  \bbg_t,\bbx_t-\bbx_{t}^* \rangle-\frac{\lambda}{2}\sum_{t=1}^T\|\bbx_{t}-\bbx_{t}^* \|^2_{{{\bbv}}_t^{1/2}}
\\ & =\sum_{t=1}^T\langle  \bbg_t,\bbx_t-\bbx_{t+1} \rangle+\sum_{t=1}^T\langle  \bbg_t,\bbx_{t+1}-\bbx_{t}^* \rangle-\frac{\lambda}{2}\sum_{t=1}^T\|\bbx_{t}-\bbx_{t}^* \|^2_{{{\bbv}}_t^{1/2}}.
\end{align}
For the second term on the RHS of \eqref{sdc}, using Lemma~\ref{lem:simple-grad-bound} by letting $A_t=\bbx_{t+1}-\bbx_{t}^*$ we have
\begin{align*}
  \langle  \bbg_t,\bbx_{t+1}-\bbx_{t}^* \rangle &\leq \frac{1}{1-\beta}\big(\langle \bbx_{t+1}-\bbx_{t}^*,\bbm_{t} \rangle-\langle \bbx_{t}-\bbx_{t-1}^*,\bbm_{t-1} \rangle\big)+\langle \bbx_{t}-\bbx_{t-1}^*,\bbm_{t-1} \rangle
\\\nonumber&+\frac{\beta}{1-\beta}\langle \bbx_{t}-\bbx_{t-1}^*-\bbx_{t+1}+\bbx_{t}^*,\bbm_{t-1} \rangle,
\end{align*}
which implies
\begin{align*}
\nonumber &\quad\sum_{t=1}^T \langle  \bbg_t,\bbx_{t+1}-\bbx_{t}^* \rangle
\\\nonumber &\leq
\frac{1}{1-\beta}\big(\langle \bbx_{T+1}-\bbx_{T}^*,\bbm_{T} \rangle-\langle \bbx_{1}-\bbx_{0}^*,\bbm_{0} \rangle\big)+\langle \bbx_{1}-\bbx_{0}^*,\bbm_{0} \rangle
+\sum_{t=1}^{T-1}\langle \bbx_{t+1}-\bbx_{t}^*,\bbm_{t} \rangle
\\\nonumber&+\frac{\beta}{1-\beta}\sum_{t=1}^T\langle \bbx_{t}-\bbx_{t+1},\bbm_{t-1} \rangle
+\frac{\beta}{1-\beta}\sum_{t=1}^T\langle \bbx_{t}^*-\bbx_{t-1}^*,\bbm_{t-1} \rangle
\\\nonumber&=\frac{\beta}{1-\beta}\langle \bbx_{T+1}-\bbx_{T}^*,\bbm_{T} \rangle
+\sum_{t=1}^{T}\langle \bbx_{t+1}-\bbx_{t}^*,\bbm_{t} \rangle
+\frac{\beta}{1-\beta}\sum_{t=1}^T\langle \bbx_{t}-\bbx_{t+1},\bbm_{t-1} \rangle
\\&+\frac{\beta}{1-\beta}\sum_{t=1}^T\langle \bbx_{t}^*-\bbx_{t-1}^*,\bbm_{t-1} \rangle.
\end{align*}
Substituting above into \eqref{sdc}, we get
\begin{align} \label{eqn:thm1:1}
\nonumber \sum_{t=1}^T \big(f_{t}(\bbx_{t}) - f_{t}(\bbx_{t}^*)\big)
&\leq \sum_{t=1}^T\langle  \bbg_t,\bbx_t-\bbx_{t+1} \rangle+
\frac{\beta}{1-\beta}\langle \bbx_{T+1}-\bbx_{T}^*,\bbm_{T} \rangle
\\\nonumber &+\sum_{t=1}^{T}\langle \bbx_{t+1}-\bbx_{t}^*,\bbm_{t} \rangle
+\frac{\beta}{1-\beta}\sum_{t=1}^T\langle \bbx_{t}-\bbx_{t+1},\bbm_{t-1} \rangle
\\&+\frac{\beta}{1-\beta}\sum_{t=1}^T\langle \bbx_{t}^*-\bbx_{t-1}^*,\bbm_{t-1} \rangle-\frac{\lambda}{2}\sum_{t=1}^T\|\bbx_{t}-\bbx_{t}^* \|^2_{{{\bbv}}_t^{1/2}}.
\end{align}
We use Lemma~\ref{lem:strong:convexsss} to derive an upper bound
for $\langle \bbx_{t+1}-\bbx_{t}^*,\bbm_{t} \rangle$ in \eqref{eqn:thm1:1} as
  \begin{align}\label{mmmx}
  \nonumber  \langle \bbm_t,\bbx_{t+1} -\bbx_{t}^* \rangle &\leq \frac{1}{2\alpha} \|\bbx_{t} -\bbx_{t}^* \|^2_{{\bbv}_t^{1/2}}
-\frac{1}{2\alpha} \|\bbx_{t+1} -\bbx_{t}^* \|^2_{{\bbv}_t^{1/2}}
    -\frac{1}{2\alpha} \|\bbx_{t+1}-\bbx_{t} \|^2_{{\bbv}_t^{1/2}}
    \\&\leq  \frac{1}{2\alpha} \|\bbx_{t} -\bbx_{t}^* \|^2_{{\bbv}_t^{1/2}}
-\frac{1}{2\alpha}\|\bbx_{t+1} -\bbx_{t}^* \|^2_{{\bbv}_t^{1/2}}-\frac{L}{2} \|\bbx_{t+1}-\bbx_{t} \|^2_{{\bbv}_{t}^{1/2}},
  \end{align}
where the second inequality holds because $1/\alpha \geq L$.
\\
Plugging \eqref{mmmx} into \eqref{eqn:thm1:1} and using $L$-smoothness assumption, we have
\begin{align} \label{eqn:thm1:2nn}
\nonumber &\quad \sum_{t=1}^T \big(f_t(\bbx_{t+1}) - f_{t}(\bbx_{t}^*)\big)
\\\nonumber &\leq
\frac{\beta}{1-\beta}\Big(\langle \bbx_{T+1}-\bbx_{T}^*,\bbm_{T} \rangle
+\sum_{t=1}^T\langle \bbx_{t}-\bbx_{t+1},\bbm_{t-1} \rangle
+\sum_{t=1}^T\langle \bbx_{t}^*-\bbx_{t-1}^*,\bbm_{t-1} \rangle\Big)
\\ &+\frac{1}{2\alpha} \sum_{t=1}^{T}\|\bbx_{t} -\bbx_{t}^* \|^2_{{\bbv}_t^{1/2}}
-\frac{1}{2\alpha}\sum_{t=1}^{T}\|\bbx_{t+1} -\bbx_{t}^* \|^2_{{\bbv}_t^{1/2}}
-\frac{\lambda}{2}\sum_{t=1}^T\|\bbx_{t}-\bbx_{t}^* \|^2_{{{\bbv}}_t^{1/2}}.
\end{align}
We use Lemma~\ref{lem:support} to show the lower
bound for LHS of \eqref{eqn:thm1:2nn} as
\begin{subequations}
\begin{equation} \label{eqn:lem1:53}
f_t(\bbx_{t+1})-f_t(\bbx_{t}^*) \geq \frac{\lambda}{2} \|\bbx_{t+1}-\bbx_{t}^*\|^2_{{{\bbv}}_{t}^{1/2}}.
\end{equation}
Next, we bound the first three terms on the RHS of \eqref{eqn:thm1:2nn}.
\begin{itemize}
  \item Bound for $\sum_{t=1}^T\langle \bbx_{t}^*-\bbx_{t-1}^*,\bbm_{t-1} \rangle$ in \eqref{eqn:thm1:2nn}.

   From the fact that $\bbm_{0} =0$, we have
  \begin{align}\label{ssss}
 \nonumber  \sum_{t=1}^T\langle \bbx_{t}^*-\bbx_{t-1}^*,\bbm_{t-1} \rangle&= \sum_{t=1}^T\sum_{i=1}^{p}\langle x_{t,i}^*-x_{t-1,i}^*,m_{t-1,i} \rangle
\\\nonumber &=\sum_{t=2}^T\sum_{i=1}^{p}\langle x_{t,i}^*-x_{t-1,i}^*,m_{t-1,i} \rangle
 \\\nonumber &= \sum_{t=2}^T\sum_{i=1}^{p}\langle {\upsilon}_{t-1,i}^{1/2}(x_{t,i}^*-x_{t-1,i}^*),{\upsilon}_{t-1,i}^{-1/2}m_{t-1,i} \rangle
     \\\nonumber &\leq \max_{i\in [p]}({\upsilon}_{t-1,i}^{-1/2}m_{t-1,i} )\sum_{t=2}^T\sum_{i=1}^{p}{\upsilon}_{t-1,i}^{1/2}| x_{t,i}^*-x_{t-1,i}^*|
     \\&\leq \beta \sum_{i=1}^{p} {\upsilon}_{T,i}^{1/2}{C}^*_{T,i},
  \end{align}
   where the last inequality follows from the fact that ${\upsilon}_{t,i}^{1/2}\geq {\upsilon}_{t-1,i}^{1/2}$, and ${C}^*_{T,i}$ is defined as in \eqref{CTT}.

\item Bound for $\langle \bbm_{T},\bbx_{T+1}-\bbx_{T}^* \rangle$ in \eqref{eqn:thm1:2nn}.

  From H\"{o}lder's inequality, we have
  \begin{align}\label{ss}
  \nonumber  \langle \bbm_{T},\bbx_{T+1}-\bbx_{T}^* \rangle &\leq \|\bbm_{T} \|_{{\bbv}_{T}^{-1/2}}\|\bbx_{T+1}-\bbx_{T}^*\|_{{\bbv}_{T}^{1/2}}
      \\\nonumber &\leq \frac{\alpha}{2}\|\bbm_{T} \|_{{\bbv}_{T}^{-1/2}}^2+\frac{1}{2\alpha}\|\bbx_{T+1}-\bbx_{T}^*\|_{{\bbv}_{T}^{1/2}}^2
      \\&\leq \frac{\alpha}{2}\|\bbm_{T} \|_{{\bbv}_{T}^{-1/2}}^2+\frac{D_{\infty}^2}{2\alpha}\sum_{i=1}^{p} {\upsilon}_{T,i}^{1/2},
  \end{align}
  where the last inequality follows from Assumption~\ref{as3}.
\item Bound for $\sum_{t=1}^T\langle \bbm_{t-1},\bbx_{t}-\bbx_{t+1} \rangle$ in \eqref{eqn:thm1:2nn}.

From H\"{o}lder's inequality, we have
  \begin{align*}
\nonumber  \langle \bbm_{t-1},\bbx_{t}-\bbx_{t+1} \rangle
 & \leq  \|\bbm_{t-1} \|_{{\bbv}_{t}^{-1/2}}\|\bbx_{t+1}-\bbx_{t} \|_{{\bbv}_{t}^{1/2}}
  \\\nonumber
  &\leq  \frac{\alpha}{2}\|\bbm_{t-1} \|^2_{{\bbv}_{t}^{-1/2}}+\frac{1}{2\alpha}\|\bbx_{t+1}-\bbx_{t} \|^2_{{{\bbv}}_{t}^{1/2}}
  \\\nonumber
  &\leq  \frac{\alpha}{2}\|\bbm_{t-1} \|^2_{{{\bbv}}_{t-1}^{-1/2}}+\frac{1}{2\alpha}\|\bbx_{t+1}-\bbx_{t} \|^2_{{{\bbv}}_{t}^{1/2}}
  \\\nonumber
  &\leq  \frac{\alpha}{2}\|\bbm_{t-1} \|^2_{{{\bbv}}_{t-1}^{-1/2}}+\frac{1}{2\alpha}\|\Pi_{\mathcal{X}, {{\bbv}}^{1/2}_t}\big(\bbx_{t} - \alpha {{\bbv}}_t^{-1/2} \bbm_t\big)-\Pi_{\mathcal{X}, {{\bbv}}^{1/2}_t}\big(\bbx_{t}\big) \|^2_{{{\bbv}}_t^{1/2}}
  \\&\leq   \frac{\alpha}{2}\|\bbm_{t-1} \|^2_{{{\bbv}}_{t-1}^{-1/2}}+\frac{\alpha}{2}\|{{\bbv}}_t^{-1/2}\bbm_{t} \|^2_{{{\bbv}}_t^{1/2}},
  \end{align*}
  where the second inequality is due to Young's inequality, the third inequality is derived from ${{\upsilon}}_{t,i}^{1/2}\geq {{\upsilon}}_{t-1,i}^{1/2}$ and the last inequality is by Lemma~\ref{mc}.
  Using $\bbm_0 = 0$, we get
  \begin{align}\label{hn}
   \nonumber \sum_{t=1}^{T} \langle \bbm_{t-1},\bbx_{t}-\bbx_{t+1} \rangle
    &\leq  \frac{\alpha}{2}\sum_{t=1}^{T}\|\bbm_{t-1} \|^2_{{{\bbv}}_{t-1}^{-1/2}}+\frac{\alpha}{2}\sum_{t=1}^{T}\|{{\bbv}}_t^{-1/2}\bbm_{t} \|^2_{{{\bbv}}_t^{1/2}}
    \\\nonumber &\leq  \frac{\alpha}{2}\sum_{t=2}^{T}\|\bbm_{t-1} \|^2_{{{\bbv}}_{t-1}^{-1/2}}+\frac{\alpha}{2}\sum_{t=1}^{T}\|{{\bbv}}_t^{-1/2}\bbm_{t} \|^2_{{{\bbv}}_t^{1/2}}
    \\&=\frac{\alpha}{2}\sum_{t=1}^{T-1}\|\bbm_{t} \|^2_{{{\bbv}}_{t}^{-1/2}}+\frac{\alpha}{2}\sum_{t=1}^{T}\|{{\bbv}}_t^{-1/2}\bbm_{t} \|^2_{{{\bbv}}_t^{1/2}}.
  \end{align}
\end{itemize}
 \end{subequations}
Plugging \eqref{eqn:lem1:53}-\eqref{hn} into \eqref{eqn:thm1:2nn} leads to
\begin{align*}
\nonumber&\quad(\frac{\lambda}{2}+\frac{1}{2\alpha}) \sum_{t=1}^{T} \sum_{i=1}^p {\upsilon}^{1/2}_{t,i}(x_{t+1,i}- x_{t,i}^*)^2
\\\nonumber &\leq
(\frac{1}{2\alpha}-\frac{\lambda}{2})\sum_{t=1}^{T} \sum_{i=1}^p {\upsilon}^{1/2}_{t,i}(x_{t,i}- x_{t,i}^*)^2
+\frac{\beta D_{\infty}^2}{2(1-\beta)\alpha}\sum_{i=1}^{p} {\upsilon}_{T,i}^{1/2} \\\nonumber
 &+\frac{\alpha\beta}{1-\beta}\sum_{t=1}^{T}\|\bbm_{t} \|_{{\bbv}_t^{-1/2}}^2
     +\frac{\beta^2 }{1-\beta}\sum_{i=1}^{p} {\upsilon}_{T,i}^{1/2}{C}^*_{T,i}
     \\\nonumber &\leq
(\frac{1}{2\alpha}-\frac{\lambda}{2})\sum_{t=1}^{T} \sum_{i=1}^p {\upsilon}^{1/2}_{t,i}(x_{t,i}- x_{t,i}^*)^2
+\frac{\beta D_{\infty}^2}{2(1-\beta)\alpha}\sum_{i=1}^{p} {\upsilon}_{T,i}^{1/2} \\
 &+\frac{2\alpha\beta}{(1 - \beta)^3} \sum_{i=1}^p \|g_{1:T,i} \|
     +\frac{\beta^2 }{1-\beta}\sum_{i=1}^{p} {\upsilon}_{T,i}^{1/2}{C}^*_{T,i},
\end{align*}
where the last inequality follows from Lemma~\ref{lem: first_lem_sadam}.

By rearranging the inequality above, we obtain:
\begin{align*}
\nonumber \sum_{t=1}^{T} \sum_{i=1}^p {\upsilon}^{1/2}_{t,i}(x_{t+1,i}- x_{t,i}^*)^2&\leq
\big(1-\frac{2\lambda}{\frac{1}{\alpha}+\lambda}\big)\sum_{t=1}^{T} \sum_{i=1}^p {\upsilon}^{1/2}_{t,i}(x_{t,i}- x_{t,i}^*)^2
+\big(\frac{2}{\lambda+\frac{1}{\alpha}}\big)\frac{\beta \vartheta}{1-\beta},
\end{align*}
where $\vartheta$ is defined as in Lemma \ref{lem:strong:convex}.
\end{proof}

\subsection*{Details of Section \ref{sec:main2}}

Using similar argument to proof of Lemma \ref{lem:strong:convexsss}, we have the following Lemma.
\begin{lem}(Counterpart of Lemma \ref{lem:strong:convexsss}).\label{lem:strong:convexsss2}
The decision sequence $\{\bbz^j_{t}\}$ generated by Algorithm~\ref{alg:subgradient2} satisfies
\begin{align*}
 \langle \bbm_t^j,\bbz^{j+1}_{t}-\bbx_t^*\rangle &\leq \frac{1}{2 \alpha} \|\bbx_t^*- \bbz^j_t\|_{({\bbv}^j_{t})^{1/2}}^2
- \frac{1}{2 \alpha} \|\bbz^{j+1}_{t}- \bbx_t^*\|_{({\bbv}^j_{t})^{1/2}}^2- \frac{1}{2 \alpha} \|\bbz^{j+1}_{t}- \bbz^j_t\|_{({\bbv}^j_{t})^{1/2}}^2,
\end{align*}
where $\bbx_t^* =\argmin_{\bbx \in \mathcal{X}} f_t(\bbx).$
\end{lem}
\subsubsection*{Proof of Lemma \ref{lem:strong:convex2}}
\begin{proof}
The steps of the proof are similar to the one for Lemma~\ref{lem:strong:convex}.
Following the lines in the proof of this Lemma, Eq. \eqref{eqn:thm1:1} will be changed to
\begin{align} \label{eqn:thm1:115}
\nonumber \sum_{t=1}^T \big(f_{t}(\bbz^{j}_{t}) - f_{t}(\bbx_{t}^*)\big)
&\leq \sum_{t=1}^T\langle  \bbg_t^j,\bbz^{j}_{t}-\bbz_{t}^{j+1} \rangle+
\frac{\beta}{1-\beta}\langle \bbz^{j+1}_{T}-\bbx_{T}^*,\bbm_{T}^j \rangle
\\\nonumber &+\sum_{t=1}^{T}\langle \bbz_{t}^{j+1}-\bbx_{t}^*,\bbm_{t}^j \rangle
+\frac{\beta}{1-\beta}\sum_{t=1}^T\langle \bbz^{j}_{t}-\bbz_{t}^{j+1},\bbm_{t-1}^j \rangle
\\&+\frac{\beta}{1-\beta}\sum_{t=1}^T\langle \bbx_{t}^*-\bbx_{t-1}^*,\bbm_{t-1}^j \rangle-\frac{\lambda}{2}\sum_{t=1}^T\|\bbz^{j}_{t}-\bbx_{t}^* \|^2_{({\bbv}^j_t)^{1/2}}.
\end{align}
We use Lemma~\ref{lem:strong:convexsss2} to derive an upper bound
for $\langle \bbz^{j+1}_{t}-\bbx_{t}^*,\bbm_{t}^j \rangle$ in \eqref{eqn:thm1:11} as
  \begin{align}\label{mmm}
  \nonumber  \langle \bbm_t^j,\bbz^{j+1}_{t} -\bbx_{t}^* \rangle &\leq \frac{1}{2\alpha} \|\bbz^{j}_{t} -\bbx_{t}^* \|^2_{({\bbv}^{j}_t)^{1/2}}
-\frac{1}{2\alpha} \|\bbz^{j+1}_{t} -\bbx_{t}^* \|^2_{({\bbv}^{j}_t)^{1/2}}
    -\frac{1}{2\alpha} \|\bbz^{j+1}_{t}-\bbz^{j}_{t} \|^2_{({\bbv}^{j}_t)^{1/2}}
    \\&\leq  \frac{1}{2\alpha} \|\bbz^{j}_{t} -\bbx_{t}^* \|^2_{({\bbv}^{j}_t)^{1/2}}
-\frac{1}{2\alpha}\|\bbz^{j+1}_{t} -\bbx_{t}^* \|^2_{({\bbv}^{j}_t)^{1/2}}-\frac{L}{2} \|\bbz^{j+1}_{t}-\bbz^{j}_{t} \|^2_{({\bbv}^{j}_t)^{1/2}},
  \end{align}
where the second inequality holds because $1/\alpha \geq L$.
\\
Plugging \eqref{mmm} into \eqref{eqn:thm1:115} and using $L$-smoothness assumption, we have
\begin{align} \label{eqn:thm1:2n}
\nonumber &\quad \sum_{t=1}^T \big(f_t(\bbz^{j+1}_{t}) - f_{t}(\bbx_{t}^*)\big)
\\\nonumber &\leq
\frac{\beta}{1-\beta}\langle \bbz^{j+1}_{T}-\bbx_{T}^*,\bbm_{T}^j \rangle
+\frac{\beta}{1-\beta}\sum_{t=1}^T\langle \bbz^{j}_{t}-\bbz^{j+1}_{t},\bbm_{t-1}^j \rangle
+\frac{\beta}{1-\beta}\sum_{t=1}^T\langle \bbx_{t}^*-\bbx_{t-1}^*,\bbm_{t-1}^j \rangle
\\ &+\frac{1}{2\alpha} \sum_{t=1}^{T}\|\bbz^{j}_{t} -\bbx_{t}^* \|^2_{({\bbv}^{j}_t)^{1/2}}
-\frac{1}{2\alpha}\sum_{t=1}^{T}\|\bbz^{j+1}_{t} -\bbx_{t}^* \|^2_{({\bbv}^{j}_t)^{1/2}}
-\frac{\lambda}{2}\sum_{t=1}^T\|\bbz^{j}_{t}-\bbx_{t}^* \|^2_{({\bbv}^{j}_t)^{1/2}}.
\end{align}
We use Lemma~\ref{lem:support} to show the lower
bound for LHS of \eqref{eqn:thm1:2n} as
\begin{subequations}
\begin{equation} \label{eqn:lem1:5324}
f_t(\bbz^{j+1}_{t})-f_t(\bbx_{t}^*) \geq \frac{\lambda}{2} \|\bbz^{j+1}_{t}-\bbx_{t}^*\|^2_{({\bbv}^{j}_t)^{1/2}}.
\end{equation}
Next, we bound the first three terms on the RHS of \eqref{eqn:thm1:2n}.
\begin{itemize}
  \item Bound for $\sum_{t=1}^T\langle \bbx_{t}^*-\bbx_{t-1}^*,\bbm_{t-1}^j \rangle$ in \eqref{eqn:thm1:2n}.

  From the fact that $\bbm^{j}_{0} =0$, we have
\begin{align*}
 \nonumber  &\quad \sum_{t=1}^T\langle \bbx_{t}^*-\bbx_{t-1}^*,\bbm^{j}_{t-1} \rangle
\\\nonumber &=\sum_{t=2}^T\sum_{i=1}^{p}\langle (\upsilon_{t-1,i}^{j})^{1/4}(x_{t,i}^*-x_{t-1,i}^*),(\upsilon_{t-1,i}^{j})^{-1/4}m^{j}_{t-1,i} \rangle\\
\nonumber &\leq
\sum_{t=2}^T\|({\bbv}^{j}_{t-1})^{1/4}( \bbx_{t}^*-\bbx_{t-1}^*)\| \|({\bbv}^{j}_{t-1})^{-1/4}\bbm^{j}_{t-1} \|
     \\\nonumber&\leq\frac{1}{2}\sum_{t=2}^T\|({\bbv}^{j}_{t-1})^{1/4}(\bbx_{t}^*-\bbx_{t-1}^*) \|^2+\frac{1}{2}\sum_{t=2}^T\|({\bbv}^{j}_{t-1})^{-1/4}\bbm^{j}_{t-1}\|^2\\
      &\leq  \frac{1}{2}\sum_{i=1}^{p}(\upsilon_{T,i}^{j})^{1/2}\sum_{t=2}^T(x_{t,i}^*-x_{t-1,i}^*)^2+\frac{1}{2}\sum_{t=2}^T\|({\bbv}^{j}_{t-1})^{-1/4}\bbm^{j}_{t-1}\|^2,
\end{align*}
  where the first inequality derives from the Cauchy-Schwarz inequality, the second inequality follows from Young's inequality
  and the last inequality holds due to the fact that $(\upsilon_{t,i}^j)^{1/2}\geq (\upsilon_{t-1,i}^j)^{1/2}$.
  Then, by definition of ${S}^*_{T,i}$ in \eqref{STT}, we have
  \begin{align}\label{ssss2}
     \sum_{t=1}^T\langle \bbx_{t}^*-\bbx_{t-1}^*,\bbm^{j}_{t-1} \rangle  &\leq \frac{1}{2}\sum_{i=1}^{p}(\upsilon_{T,i}^{j})^{1/2}{S}^*_{T,i}+\frac{1}{2}\sum_{t=1}^{T} \| \bbm_{t}\|_{(\bbv_t^j)^{-1/2}} ^ 2.
  \end{align}

\item Bound for $\langle \bbm_{T},\bbz^{j+1}_{T}-\bbx_{T}^* \rangle$ in \eqref{eqn:thm1:2n}.

With the same argument as in the proof of \eqref{ss}, we obtain
  \begin{align}\label{ssF}
  \langle \bbm_{T}^j,\bbz^{j+1}-\bbx_{T}^* \rangle &\leq \frac{\alpha}{2}\|\bbm_{T}^j \|_{({\bbv}_{T}^j)^{-1/2}}^2+\frac{D_{\infty}^2}{2\alpha}\sum_{i=1}^{p} ({\upsilon}_{T,i}^j)^{1/2}.
  \end{align}

\item Bound for $\sum_{t=1}^T\langle \bbz^{j}_{t}-\bbz^{j+1}_{t},\bbm_{t-1}^j \rangle$ in \eqref{eqn:thm1:2n}.

We use the same lines of argument for \eqref{hn} to acquire:
  \begin{align}\label{hnc}
 \sum_{t=1}^T\langle \bbz^{j}_{t}-\bbz^{j+1}_{t},\bbm_{t-1}^j \rangle
    &\leq  \frac{\alpha}{2}\sum_{t=1}^{T-1}\|\bbm_{t}^j \|^2_{({{\bbv}}_{t}^j)^{-1/2}}
    +\frac{\alpha}{2}\sum_{t=1}^{T}\|({{\bbv}}_t^j)^{-1/2}\bbm_{t}^j \|^2_{({{\bbv}}_t^j)^{1/2}}.
  \end{align}
\end{itemize}
 \end{subequations}
Plugging \eqref{eqn:lem1:5324}-\eqref{hnc} into \eqref{eqn:thm1:2n} leads to
\begin{align*}
\nonumber&\quad(\frac{\lambda}{2}+\frac{1}{2\alpha}) \sum_{t=1}^{T} \sum_{i=1}^p ({\upsilon}_{t,i}^j)^{1/2}(z_{t,i}^{j+1}- x_{t,i}^*)^2
\\\nonumber &\leq
(\frac{1}{2\alpha}-\frac{\lambda}{2})\sum_{t=1}^{T} \sum_{i=1}^p ({\upsilon}_{t,i}^j)^{1/2}(z_{t,i}^j- x_{t,i}^*)^2
+\frac{\beta D_{\infty}^2}{2(1-\beta)\alpha}\sum_{i=1}^{p} ({\upsilon}_{T,i}^j)^{1/2} \\\nonumber
 &+\frac{\alpha\beta}{1-\beta}\sum_{t=1}^{T}\|\bbm_{t}^j \|_{({\bbv}_t^j)^{-1/2}}^2
     +\frac{\beta}{2(1-\beta)}\sum_{i=1}^{p}(\upsilon_{T,i}^{j})^{1/2}S^*_{T,i}
     \\\nonumber &\leq
(\frac{1}{2\alpha}-\frac{\lambda}{2})\sum_{t=1}^{T} \sum_{i=1}^p ({\upsilon}_{t,i}^j)^{1/2}(z_{t,i}^j- x_{t,i}^*)^2
+\frac{\beta D_{\infty}^2}{2(1-\beta)\alpha}\sum_{i=1}^{p} ({\upsilon}_{T,i}^j)^{1/2} \\
 &+\frac{2\alpha\beta}{(1 - \beta)^3} \sum_{i=1}^p \|g_{1:T,i}^j \|
     +\frac{\beta}{2(1-\beta)}\sum_{i=1}^{p}(\upsilon_{T,i}^{j})^{1/2}S^*_{T,i},
\end{align*}
where the last inequality follows from Lemma~\ref{lem: first_lem_sadam}.

By rearranging the inequality above, we obtain:
\begin{align*}
\nonumber &\quad \sum_{t=1}^{T} \sum_{i=1}^p {({\upsilon}^{j}_{t,i})^{1/2}}(z^{j+1}_{t,i}- x_{t,i}^*)^2
\\&\leq
\big(1-\frac{2\lambda}{\frac{1}{\alpha}+\lambda}\big)\sum_{t=1}^{T} \sum_{i=1}^p {({\upsilon}^{j}_{t,i})^{1/2}}(z^j_{t,i}- x_{t,i}^*)^2
+\big(\frac{2}{\lambda+\frac{1}{\alpha}}\big)(\frac{\beta \vartheta}{1-\beta}),
\end{align*}
where $\vartheta$ is defined as in Lemma \ref{lem:strong:convex2}.
\end{proof}
\bibliography{ref}

\begin{thebibliography}{10}

\bibitem{hazan2016introduction}
E.~Hazan {\em et~al.}, ``Introduction to online convex optimization,'' {\em
  Foundations and Trends{\textregistered} in Optimization}, vol.~2, no.~3-4,
  pp.~157--325, 2016.

\bibitem{hosseini2016online}
S.~Hosseini, A.~Chapman, and M.~Mesbahi, ``Online distributed convex
  optimization on dynamic networks,'' {\em IEEE Transactions on Automatic
  Control}, vol.~61, no.~11, pp.~3545--3550, 2016.

\bibitem{zinkevich2003online}
M.~Zinkevich, ``Online convex programming and generalized infinitesimal
  gradient ascent,'' in {\em Proceedings of the 20th international conference
  on machine learning (icml-03)}, pp.~928--936, 2003.

\bibitem{hazan2007logarithmic}
E.~Hazan, A.~Agarwal, and S.~Kale, ``Logarithmic regret algorithms for online
  convex optimization,'' {\em Machine Learning}, vol.~69, no.~2-3,
  pp.~169--192, 2007.

\bibitem{jadbabaie2015online}
A.~Jadbabaie, A.~Rakhlin, S.~Shahrampour, and K.~Sridharan, ``Online
  optimization: Competing with dynamic comparators.,'' in {\em AISTATS}, 2015.

\bibitem{besbes2015non}
O.~Besbes, Y.~Gur, and A.~Zeevi, ``Non-stationary stochastic optimization,''
  {\em Operations Research}, vol.~63, no.~5, pp.~1227--1244, 2015.

\bibitem{yang2016tracking}
T.~Yang, L.~Zhang, R.~Jin, and J.~Yi, ``Tracking slowly moving clairvoyant:
  Optimal dynamic regret of online learning with true and noisy gradient,'' in
  {\em International Conference on Machine Learning}, pp.~449--457, PMLR, 2016.

\bibitem{mokhtari2016online}
A.~Mokhtari, S.~Shahrampour, A.~Jadbabaie, and A.~Ribeiro, ``Online
  optimization in dynamic environments: Improved regret rates for strongly
  convex problems,'' in {\em Decision and Control (CDC), 2016 IEEE 55th
  Conference on}, pp.~7195--7201, IEEE, 2016.

\bibitem{zhang2017improved}
L.~Zhang, T.~Yang, J.~Yi, J.~Rong, and Z.-H. Zhou, ``Improved dynamic regret
  for non-degenerate functions,'' in {\em Advances in Neural Information
  Processing Systems}, pp.~732--741, 2017.

\bibitem{nazari2021dynamic}
P.~Nazari and E.~Khorram, ``Dynamic regret analysis for online meta-learning,''
  {\em arXiv preprint arXiv:2109.14375}, 2021.

\bibitem{hall2013dynamical}
E.~C. Hall and R.~M. Willett, ``Dynamical models and tracking regret in online
  convex programming,'' {\em arXiv preprint arXiv:1301.1254}, 2013.

\bibitem{chiang2012online}
C.-K. Chiang, T.~Yang, C.-J. Lee, M.~Mahdavi, C.-J. Lu, R.~Jin, and S.~Zhu,
  ``Online optimization with gradual variations,'' in {\em Conference on
  Learning Theory}, pp.~6--1, JMLR Workshop and Conference Proceedings, 2012.

\bibitem{duchi2011adaptive}
J.~Duchi, E.~Hazan, and Y.~Singer, ``Adaptive subgradient methods for online
  learning and stochastic optimization.,'' {\em Journal of machine learning
  research}, vol.~12, no.~7, 2011.

\bibitem{kingma2014adam}
D.~P. Kingma and J.~Ba, ``Adam: A method for stochastic optimization,'' {\em
  arXiv preprint arXiv:1412.6980}, 2014.

\bibitem{reddi2019convergence}
S.~J. Reddi, S.~Kale, and S.~Kumar, ``On the convergence of adam and beyond,''
  {\em arXiv preprint arXiv:1904.09237}, 2019.

\bibitem{chen2018sadagrad}
Z.~Chen, Y.~Xu, E.~Chen, and T.~Yang, ``Sadagrad: Strongly adaptive stochastic
  gradient methods,'' in {\em International Conference on Machine Learning},
  pp.~913--921, PMLR, 2018.

\bibitem{wang2019sadam}
G.~Wang, S.~Lu, W.~Tu, and L.~Zhang, ``Sadam: A variant of adam for strongly
  convex functions,'' {\em arXiv preprint arXiv:1905.02957}, 2019.

\bibitem{nazari2020adaptive}
P.~Nazari, D.~A. Tarzanagh, and G.~Michailidis, ``Adaptive first-and
  zeroth-order methods for weakly convex stochastic optimization problems,''
  {\em arXiv preprint arXiv:2005.09261}, 2020.

\bibitem{mukkamala2017variants}
M.~C. Mukkamala and M.~Hein, ``Variants of rmsprop and adagrad with logarithmic
  regret bounds,'' in {\em International Conference on Machine Learning},
  pp.~2545--2553, PMLR, 2017.

\bibitem{nazari2019dadam}
P.~Nazari, D.~A. Tarzanagh, and G.~Michailidis, ``Dadam: A consensus-based
  distributed adaptive gradient method for online optimization,'' {\em arXiv
  preprint arXiv:1901.09109}, 2019.

\bibitem{tieleman2017divide}
T.~Tieleman and G.~Hinton, ``Divide the gradient by a running average of its
  recent magnitude. coursera: Neural networks for machine learning,'' {\em
  Technical Report}, 2017.

\bibitem{zeiler2012adadelta}
M.~D. Zeiler, ``Adadelta: an adaptive learning rate method,'' {\em arXiv
  preprint arXiv:1212.5701}, 2012.

\bibitem{hazan2019introduction}
E.~Hazan, ``Introduction to online convex optimization,'' {\em arXiv preprint
  arXiv:1909.05207}, 2019.

\bibitem{van2016metagrad}
T.~van Erven and W.~M. Koolen, ``Metagrad: Multiple learning rates in online
  learning,'' {\em arXiv preprint arXiv:1604.08740}, 2016.

\bibitem{alacaoglu2020new}
A.~Alacaoglu, Y.~Malitsky, P.~Mertikopoulos, and V.~Cevher, ``A new regret
  analysis for adam-type algorithms,'' {\em arXiv preprint arXiv:2003.09729},
  2020.

\bibitem{reddi2018convergence}
S.~J. Reddi, S.~Kale, and S.~Kumar, ``On the convergence of adam and beyond,''
  in {\em International Conference on Learning Representations}, 2018.

\bibitem{nazari2019adaptive}
P.~Nazari, E.~Khorram, and D.~A. Tarzanagh, ``Adaptive online distributed
  optimization in dynamic environments,'' {\em Optimization Methods and
  Software}, pp.~1--25, 2019.

\bibitem{hazan2011beyond}
E.~Hazan and S.~Kale, ``Beyond the regret minimization barrier: an optimal
  algorithm for stochastic strongly-convex optimization,'' in {\em Proceedings
  of the 24th Annual Conference on Learning Theory}, pp.~421--436, 2011.

\bibitem{mcmahan2010adaptive}
H.~B. McMahan and M.~Streeter, ``Adaptive bound optimization for online convex
  optimization,'' {\em arXiv preprint arXiv:1002.4908}, 2010.

\bibitem{auer2002adaptive}
P.~Auer, N.~Cesa-Bianchi, and C.~Gentile, ``Adaptive and self-confident on-line
  learning algorithms,'' {\em Journal of Computer and System Sciences},
  vol.~64, no.~1, pp.~48--75, 2002.

\end{thebibliography}
\bibliographystyle{ieeetr}
\end{document}